\documentclass{article} % For LaTeX2e
\usepackage{iclr2025_conference,times}

\usepackage[utf8]{inputenc} % allow utf-8 input
\usepackage[T1]{fontenc}    % use 8-bit T1 fonts
\usepackage{hyperref}       % hyperlinks
\usepackage{url}            % simple URL typesetting
\usepackage{booktabs}       % professional-quality tables
\usepackage{amsfonts}       % blackboard math symbols
\usepackage{nicefrac}       % compact symbols for 1/2, etc.
\usepackage{microtype}      % microtypography
\usepackage{xcolor}         % colors
\usepackage{graphicx}
\usepackage{wrapfig}
\usepackage{graphicx}
\usepackage[strict]{changepage}
\usepackage{amsmath}

\iclrfinalcopy % Uncomment for camera-ready version, but NOT for submission.

\title{SEED-Story: Multimodal Long Story Generation with Large Language Model}

\author{%
Shuai Yang~$^{1}$\thanks{Worked done at ARC Lab, Tencent PCG}\qquad
Yuying Ge~$^{2}$\thanks{Corresponding Authors}\qquad
Yang Li~$^{1}$\qquad
Yukang Chen~$^{4}$\qquad
Yixiao Ge~$^{2,3}$\qquad
\\[0.1cm]
\textbf{Ying Shan~$^{2,3}$\qquad
Yingcong Chen~$^{1,5\dag}$}
\\[0.2cm]$^1$HKUST(GZ)\qquad
$^2$ARC Lab, Tencent PCG\qquad
$^3$Tencent AI Lab\qquad
$^4$CUHK\qquad
$^5$HKUST\qquad
}

\begin{document}

\maketitle

\vspace{-0.2in}
\begin{figure}[h!]
\centering
\includegraphics[width=1.0\linewidth]{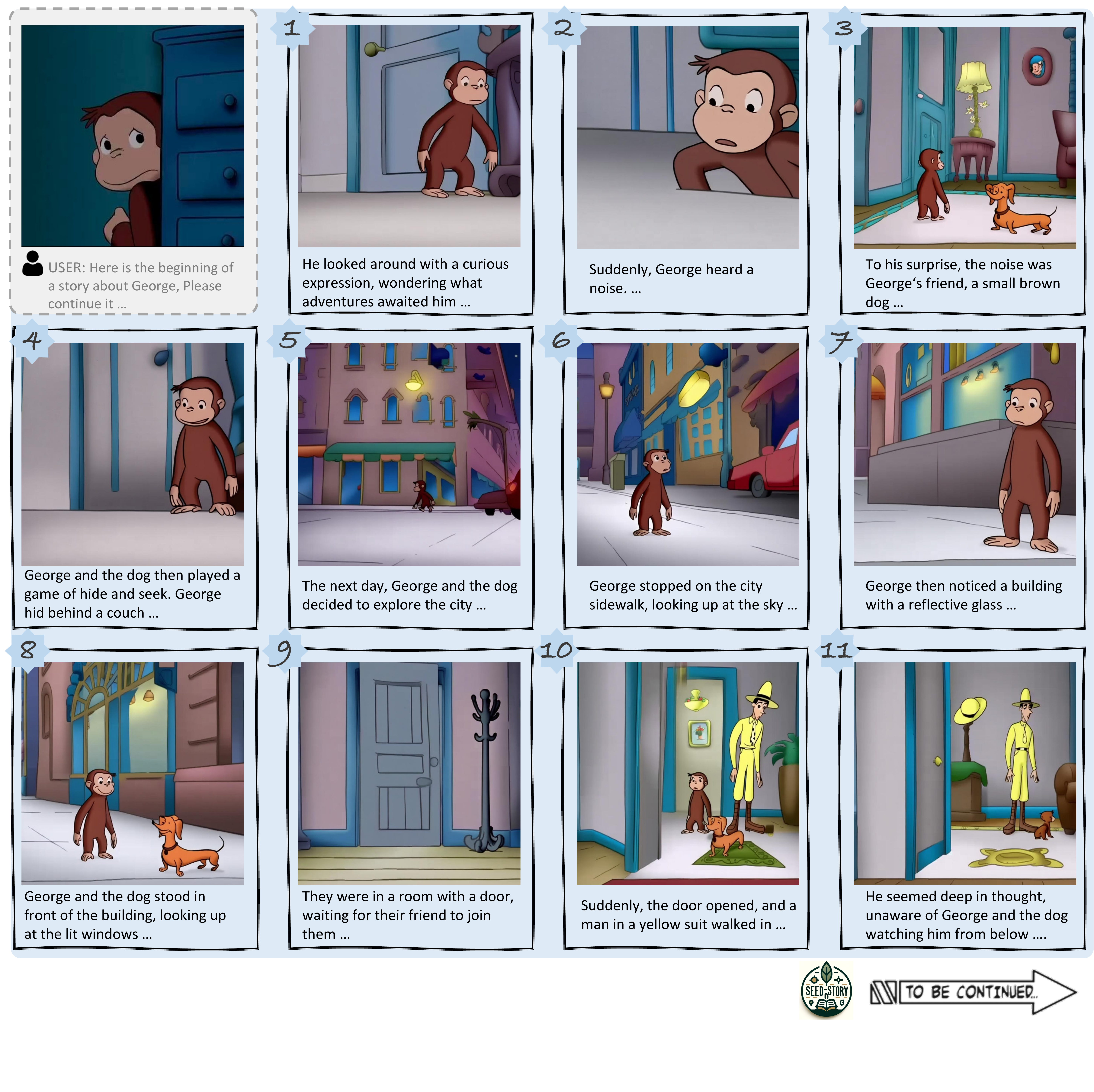}
% \vspace{0.1in}
\vspace{-35pt}
\caption{The introduced SEED-Story, powered by MLLM, is capable of generating \textbf{multimodal long stories} based on user-provided images and texts as the story's beginning. The generated story features rich and coherent narrative texts, accompanied by images that maintain consistency in characters and style. The story can span up to 25 multimodal sequences (see Appendix), even though we only use a maximum of 10 sequences for training.
}
\vspace{10pt}
\label{fig:teaser}
\end{figure}

\begin{abstract}
With the remarkable advancements in image generation and open-form text generation, the creation of interleaved image-text content has become an increasingly intriguing field. Multimodal story generation, characterized by producing narrative texts and vivid images in an interleaved manner, has emerged as a valuable and practical task with broad applications. However, this task poses significant challenges, as it necessitates the comprehension of the complex interplay between texts and images, and the ability to generate long sequences of coherent, contextually relevant texts and visuals. In this work, we propose SEED-Story, a novel method that leverages a Multimodal Large Language Model (MLLM) to generate extended multimodal stories. Our model, built upon the powerful comprehension capability of MLLM, predicts text tokens as well as visual tokens, which are subsequently processed with an adapted visual de-tokenizer to produce images with consistent characters and styles. We further propose multimodal attention sink mechanism to enable the generation of stories with up to 25 sequences (only 10 for training) in a highly efficient autoregressive manner. Additionally, we present a large-scale and high-resolution dataset named StoryStream for training our model and quantitatively evaluating the task of multimodal story generation in various aspects. 
All models, training and inference codes are released in \url{https://github.com/TencentARC/SEED-Story}.
%The models, codes and datasets are released in \url{https://github.com/TencentARC/SEED-Story}. 
%We anticipate that our work can promote the development of visual storytelling and further multimodal content generation.

\end{abstract}

\section{Introduction}

% image-text interleaved data comprehension and generation is important
Interleaved image-text data is ubiquitous on the internet, characterized by multiple images interspersed with text. In recent years, there has been a surge of interest in generating interleaved image-text content~\citet{tian2024mm,ge2024seedx,aiello2023jointly,dong2023dreamllm,team2024chameleon}, driven by the remarkable advancements in image generation~\citet{rombach2022high, lin2023sphinx, chen2023pixartalpha, Patashnik_2021_ICCV, wang2023steps} and open-form text generation~\cite{touvron2023llama, alpaca, zheng2023judging}. This has given rise to \textbf{Multimodal Story Generation}, an intriguing and valuable task that involves the generation of a sequence of consecutive
storylines along with their corresponding vivid images in an interleaved manner, similar to that of a serialized comic.

Different from \textbf{Personalized Story Visualization}~\cite{ruiz2023dreambooth, avrahami2024chosen, tewel2024training}, which aims to generate consistent images based on the provided captions following the pattern of text-to-image generation, multimodal story generation poses a more significant challenge due to the complexity of the inputs and the high demands of the outputs. Firstly, this task necessitates a thorough comprehension of interleaved data, where text is not only abstract and narrative in nature, but also deeply intertwined with complex images. The model must be adept at deciphering the intricate relationships between images and texts to maintain a coherent narrative flow. Secondly, this task requires the generation of not only a plausible text plot, but also visually captivating images that are consistent in characters and styles. The model should be capable of achieving coherence in the generation of both text and visuals, ensuring an engaging storytelling output.

Recently, Multimodal Large Language Models (MLLMs)~\cite{li2023blip2, zhu2023minigpt4, peng2023kosmos, bai2023qwen, liu2023llava1.5, zhang2023internlm, lin2023sphinx, laurenccon2024obelics} have showcased powerful comprehension abilities in understanding multimodal data, which makes them ideally suited for interleaved image-text content in multimodal stories. Consequently, we introduce SEED-Story, as shown in Figure~\ref{fig:teaser}, a novel approach that builds upon the MLLM to harness its comprehension strength, while further equipping it with the capability to generate coherent images align with the narrative texts.

%In the rapidly evolving domain of large language models (LLMs)~\cite{touvron2023llama, brown2020language, chowdhery2022palm} and multimodal large language models (MLLMs)~\cite{li2023blip2, zhu2023minigpt4, liu2023visual}, their robust integration capabilities make them ideal for tasks such as story generation. Inspired by these advancements, we have developed SEED-story, a specialized model that leverages the strengths of MLLMs to generate intricate text and image content.

% 所以，We proposed SEED-Story, 能让LLM生成高质量图,。

Specifically, following previous work~\cite{Emu2, ge2024seedx}, we utilize a pre-trained image tokenizer and de-tokenizer, which can decode realistic images with SD-XL~\cite{podell2023sdxl} by taking the features of a pre-trained ViT as input. During training, given the interleaved visual and textual data, we adopt the next-word prediction and image feature regression training objectives to regularize multimodal generation. A fixed number of learnable queries are fed into the MLLM, where the output hidden states are trained to reconstruct the ViT features of the target images. To further enhance the consistency of characters and styles in generated images, we propose de-tokenizer adaptation, where the regressed image features from the MLLM are fed into the de-tokenizer for tuning SD-XL.  This adaptation allows for better maintenance of coherence in low-level image details from the de-tokenizer, ensuring a more visually consistent storytelling output.

Furthermore, to enable the efficient generation of coherent long stories, we propose a multimodal attention sink mechanism based on window attention~\cite{beltagy2020longformer}, which maintains a fixed-size sliding window on the Key-Value (KV) states of the most recent tokens, as well as the beginning of text tokens, images tokens, and the end of image tokens. We empirically find that retaining these tokens will largely mitigate the model's failure with window attention when the token length surpasses the cache size, allowing our model to generalize to longer sequences than the training sequence length in an efficient manner. Our model with the proposed multimodal attention sink mechanism can generate long stories with up to 30 multimodal sequences, featuring rich text plots and diverse visual scenarios.

%Moreover, we have enhanced SEED-story to effectively manage extended narrative sequences. By incorporating an image attention sink technique, our model consistently maintains high image quality throughout longer stories. This technique allows the model to focus on relevant visual elements over extended sequences, promoting sustained narrative engagement and deepening the storytelling experience. The capability to generate longer stories enriches the narrative with more abundant plots and diverse scenarios, making the generated content more fun and engaging. 

% SEED相比diffusion based的有好处
%Compared to conventional storytelling models that rely primarily on GAN or diffusion techniques, SEED-story offers distinct advantages. Firstly, SEED-story demonstrates superior comprehension of text-image interleaved data, enabling it to solely learn from abstract and narrative text without the need for additional supervisions. 
%Secondly, SEED-story has the capability to generate textual narrative, enhancing the multimodal storytelling experience. Lastly, thanks to its image tokenization and image attention sink techniques, SEED-story efficiently handles and generate extended narrative sequences. This contrasts with diffusion-based autoregressive methods that maintain large intermediate features for long story generation.

% dataset
Additionally, we introduce a dataset named StoryStream for training and evaluating multimodal story generation. We design an automatic pipeline that leverages MLLMs to obtain a large-scale and high-resolution dataset featuring a sequence of narrative-rich texts and intriguing images, derived from animated videos. StoryStream is four times larger in terms of data volume compared to the existing largest story dataset~\cite{liu2023intelligent}, and it boasts higher image resolution, longer sequence lengths, and more detailed story narratives. We further meticulously design evaluation metrics to assess multimodal story generation, taking into account image style consistency, story engagement, and image-text coherence. The evaluation results demonstrate that our model, SEED-Story, achieves superior performance in these aspects. 

In summary, Our contributions are three-fold. (1) We propose SEED-Story, a novel method that leverages an MLLM to generate multimodal stories with rich narrative text and contextually relevant images. (2) We propose multimodal attention sink to enable the efficient generation of long stories with sequence lengths larger than those used during training. (3) We introduce StoryStream, a large-scale dataset specifically designed for training and benchmarking multimodal story generation.

%In summary, our contributions in this work are three-fold. (1) We propose SEED-Story, a novel method that leverages an MLLM to generate multimodal stories with narrative-rich text and coherent images, effectively addressing the challenges posed by this task. (2) We introduce a multimodal attention sink mechanism, enabling the efficient generation of coherent long stories with sequence lengths larger than those encountered during training. (3) We present StoryStream, a large-scale dataset specifically designed for training and benchmarking multimodal story generation, providing a valuable resource for future research in this area. Together, these contributions pave the way for advancements in multimodal storytelling and content generation

%To optimize SEED-story's capability to generate compelling narratives, we have introduced a dataset named StoryStream. Sourced from animated videos and processed using the state-of-the-art MLLMs, StoryStream provides a large-scale, high-resolution dataset featuring richly interleaved story elements. It includes long, intricate plots and detailed narrative text, making it an ideal resource for training and evaluating models on this complex task. Additionally, StoryStream serves as a benchmark, allowing us to systematically assess model performance in handling sophisticated multimodal story generation.

% contribution
% SEED-Story model to comprehend complex image and text, and producing 
% We propose image attn sink to help MLLM generating long story
% We introduce a new dataset called StoryStream. It make the model 

\section{Related Work}

\begin{figure}[h]
\centering
\includegraphics[width=0.9\linewidth]{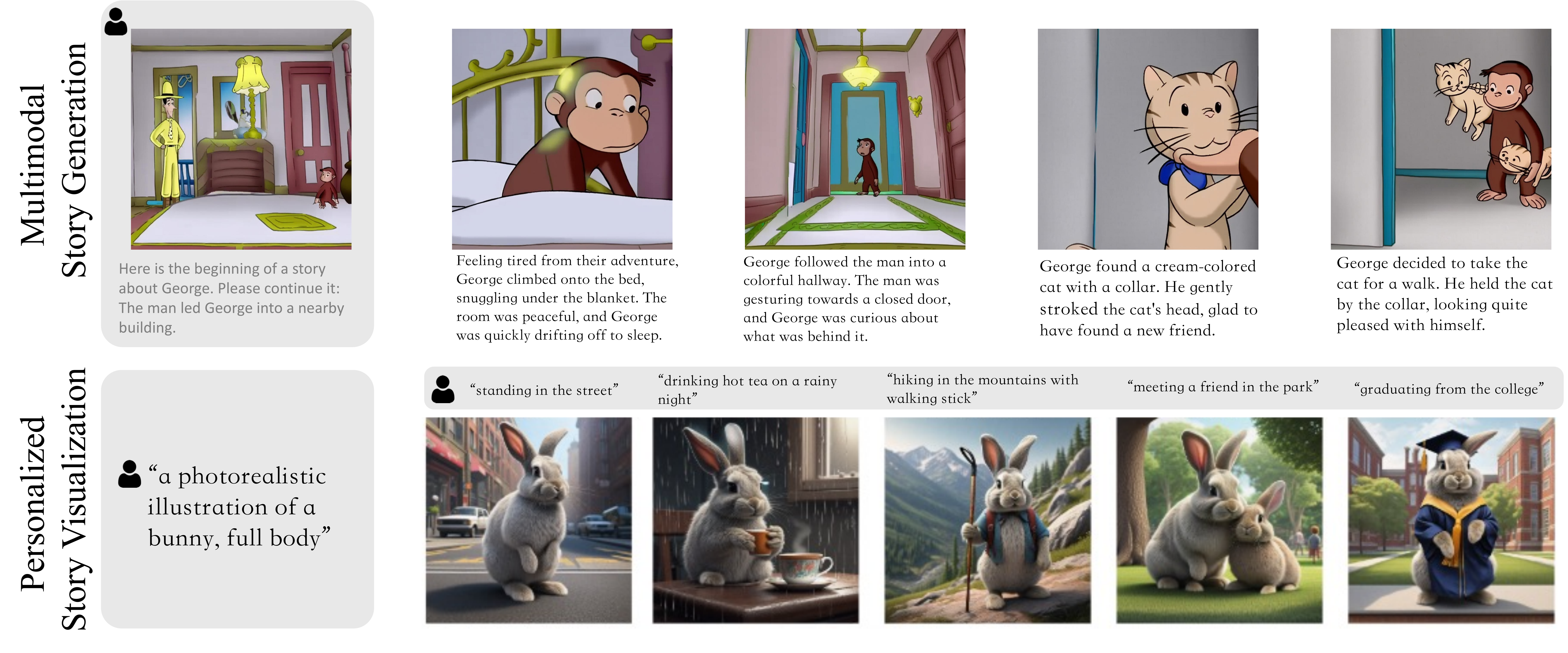}
\caption{Comparison between personalized story visualization and multimodal story generation. Contents in grey boxes refer to user's input. In the former, a sequence of captions are given (referred to as a ``story'') for consistent text-to-image generation. By contrast, multimodal story generation involves creating a sequence of consecutive storylines along with their corresponding images.}
\label{fig:personal_vs_multigen}
\end{figure}

\paragraph{Personalized Story Visualization v.s. Multimodal Story Generation}
Personalized Story Visualization, such as~\cite{li2019storygan, maharana2021improving, maharana2021integrating, maharana2022storydall, pan2024synthesizing, rahman2023make, gong2023talecrafter, liu2023intelligent, ruiz2023dreambooth, avrahami2024chosen, tewel2024training}, aims to generate images depicting specified characters engaged in various actions or within different scenes, based on \textbf{the provided captions} (so-called "story") as shown in Fig.~\ref{fig:personal_vs_multigen}, which follows the pattern of \textbf{text-to-image generation}. For example, StoryDALL-E \cite{maharana2022storydall} utilizes pre-trained models augmented with cross-attention layers to support story progression from an initial image. Innovations like AR-LDM \cite{pan2024synthesizing} and Story-LDM \cite{rahman2023make} have introduced auto-regressive diffusion models to create coherent sequences of images, while TaleCrafter \cite{gong2023talecrafter} has employed LoRA and optimization techniques to ensure consistent characters throughout complex visual narratives.
%Due to the direct provision of storylines, the need for narrative consistency is eliminated. Consequently, the resulting images are not dependent on preceding outcomes. Instead, these works generate images in parallel within a single batch. This enables the imposition of strict constraints on the cross-attention map to ensure consistency across multiple images featuring the main characters, thereby facilitating generalization to novel characters. 

Multimodal Story Generation (\cite{shen2023storygptv}) aims to generate \textbf{a sequence of consecutive storylines along with their corresponding images}, similar to that of a serialized comic. To achieve this, a model must be capable of predicting reasonable story developments and generating corresponding illustrations, by incorporating the previous results as context in an auto-regressive manner. As shown in Figure~\ref{fig:personal_vs_multigen}, the generated story images exhibit rapidly changing backgrounds and characters, different from personalized story visualization where the main character consistently appears in the images. 

The task of multimodal story generation presents a more substantial challenge, and we follow previous research~\cite{shen2023storygptv} to adopt a closed-set setting. We believe the ultimate goal of multimodal story generation should be to generate highly diverse scenarios while also generalizing to unseen characters, which will be explored in our future work.

\paragraph{MLLM for Multimodal Story Generation}
In the rapidly evolving domain of large language models (LLMs)~\cite{touvron2023llama, brown2020language, chowdhery2022palm} and multimodal large language models (MLLMs)~\cite{li2023blip2, zhu2023minigpt4, liu2023visual, peng2023kosmos2, bai2023qwenvl, liu2023improved, zhang2023internlm, lin2023sphinx, sun2023generative, yu2023scaling, ge2023making, ge2023planting, wu2023nextgpt, dong2023dreamllm, zhu2023vlgpt, Emu, li2024minigemini}, recent work StoryGPTV \cite{shen2023storygptv} explores using MLLMs for story generation by converting visual features into token embeddings for image generation, but requires aditional character and object masks for training. MM-interleaved~\cite{tian2024mm} designs a multi-scale and multi-image feature synchronizer module (MMFS) to process interleaved text-image data and generates next image conditioned on the previous context features from LLM, which makes it difficult to generate long multimodal stories due to the complex multi-scale attention mechanism.
%such as StoryGPTV \cite{shen2023storygptv}, explores using MLLMs for story generation by converting visual features into token embeddings. StoryGPTV requires character and object masks for training its LDM. In contrast, our model, with the visual tokenization and de-tokenization, have a stronger visual comprehension ability, making it trains directly from interleaved text and image data without the need for masks. MM-interleaved~\cite{tian2024mm} designs a multi-scale and multi-imagefeature synchronizer module (MMFS) to process interleaved text-image data and achieves multimodal story generation through passing LLM output features into the image decoder as the generation conditions.

%image features and LLM output. %By contrast, our approach directly generates images through LLM output tokens. %We take MM-interleaved as an important baseline and compare with it.

\begin{table}[t]
\centering
\caption{Comparison of multimodal story generation datasets. The table provides details on the number of images, their resolution, the total length of visual stories, and the average text length per sentence, which indicates the narrative detail of the text. Note that StorySalon has various size of images and we choose one of the typical sizes presented here.}
\begin{tabular}{lccccc}
\hline
\textbf{Datasets} & \textbf{Number of} & \textbf{Resolution} & \textbf{Story} & \textbf{Avg Text} \\ 
                  & \textbf{Images}   &                    & \textbf{Length} & \textbf{Length} \\ \hline
Flintstones~\citet{gupta2018imagine} & 122,560 & $128\times128$ & 5 & 86  \\
Pororo~\citet{li2019storygan} & 73,665 & $128\times128$ & 5 & 74  \\ 
StorySalon~\citet{liu2023intelligent} & 159,778 & $432\times803$ & 14 & 106  \\ 
\textbf{StoryStream} & 257,850 & $480\times854$ & 30 & 146  \\ \hline
\end{tabular}

\label{tab:visual_story_datasets}
\end{table}

\paragraph{Visual Story Dataset}
In the landscape of datasets for visual storytelling, various collections have been developed as shown in Table~\ref{tab:visual_story_datasets}. The VIST~\cite{huang2016visual} dataset is noteworthy for its use of realistic images, though it struggles with maintaining character consistency across stories. Pororo~\cite{li2019storygan} and Flintstones~\cite{gupta2018imagine} datasets, while popular for animation-based story datasets, are hindered by their low resolution and the simplicity of their accompanying texts. Another significant dataset is StorySalon~\cite{liu2023intelligent}, which offers high-resolution images and is large in scale, but it lacks global consistency across images. To address these gaps, we introduce StoryStream, a globally consistent, large-scale, high-resolution animated style dataset with engaging, narrative-rich text for complex storytelling, overcoming the limitations of existing datasets. %The analysis is shown in Table~\ref{tab:visual_story_datasets}.
%\vspace{-0.3in}
\begin{figure}[h]
\centering
\includegraphics[width=1.0\linewidth]{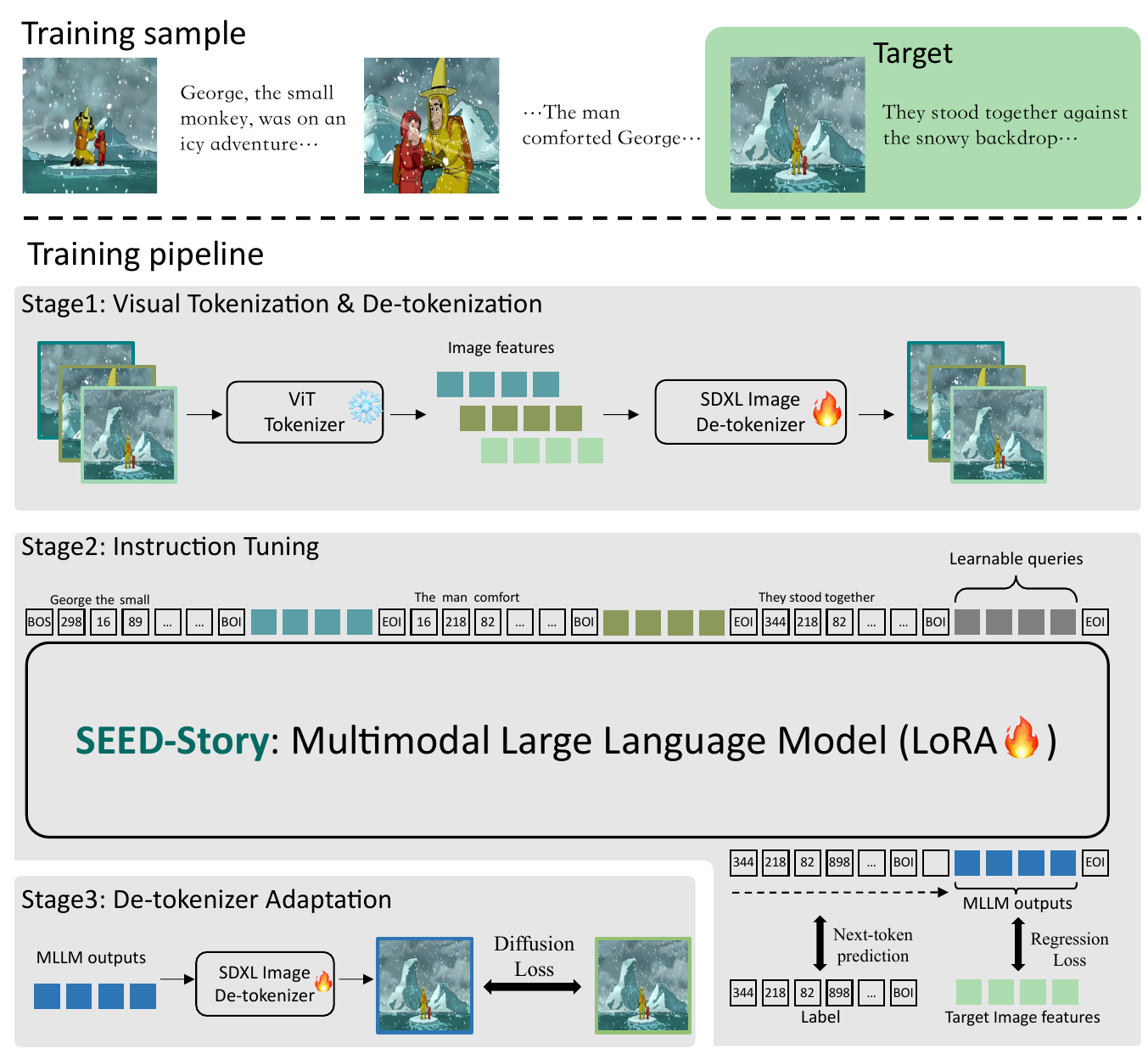}
\caption{Overview of the SEED-Story Training Pipeline: In \textbf{Stage 1}, we pre-trains an SD-XL-based de-tokenizer to reconstruct images by taking the features of a pre-trained ViT as inputs. In \textbf{Stage 2}, we sample an interleaved image-text sequence of a random length and train the MLLM by performing next-word prediction and image feature regression between the output hidden states of the learnable queries and ViT features of the target image. In \textbf{Stage 3}, the regressed image features from the MLLM are fed into the de-tokenizer for tuning SD-XL, enhancing the consistency of the characters and styles in the generated images.}
%visual tokenizer and de-tokenizer that encodes images into high-level features. We train the SDXL image decoder to reconstruct images using the image features generated by a frozen ViT; In \textbf{Stage 2}, we sample an image-text interleaved story of a random length and direct our MLLM to predict the next image and accompanying text. The regression loss is computed between the output hidden states (illustrated as dark blue squares) and the target image features (illustrated as light green squares); \textbf{Stage 3} focuses on refining the de-tokenizer to better align the LLM output space with the image feature space, enhancing the consistency of the generated image styles.}
\label{fig:pipeline}
\end{figure}

\section{Method}
% training
\subsection{Story Generation with Multimodal Large Language Model}
\paragraph{Visual Tokenization and De-tokenization}
The overview of our method is presented in Figure~\ref{fig:pipeline}.
To effectively extend visual stories, our model must comprehend and generate both images and text. Drawing inspiration from recent advancements in generative MLLMs that unify image comprehension and generation~\cite{podell2023sdxl}, we develop a multimodal story generation model.
Our model employs a pre-trained Vision Transformer~\cite{dosovitskiy2020image} (ViT) as the visual tokenizer and a pre-trained diffusion model as the visual de-tokenizer to decode images by using ViT's features as inputs. Specifically, visual embeddings from the ViT tokenizer are fed into a learnable module, which then serves as inputs for the U-Net of the pre-trained SD-XL~\cite{podell2023sdxl}. This process replaces the original text features with visual embeddings. During this stage, the parameters are optimized using open-world text-image pair data as well as story data to enhance the model's encoding-decoding capability. After this training phase, we expect the visual tokenizer and de-tokenizer modules to preserve as much image information as possible in the feature space.

%\vspace{-0.1in}
\paragraph{Story Instruction Tuning}
% training data
In our instruction tuning process for story generation, we sample a random-length subset of a story data point for each iteration. The model is tasked with predicting the next image and the next sentence of the story text. Within MLLM, all images are converted into image features using a pre-trained ViT tokenizer. 
For the target text tokens, we perform next-token prediction and use Cross Entropy loss to train for this discrete target.
For target image features, the model uses a series of learnable queries as inputs and continuously outputs a series of latents. We then compute the cosine similarity loss between the MLLM's output and the target image features. During this stage, we fine-tune the SEED-Story model using a LoRA~\citet{hu2021lora} module.

\begin{figure}[h]
\centering
\includegraphics[width=1.0\linewidth]{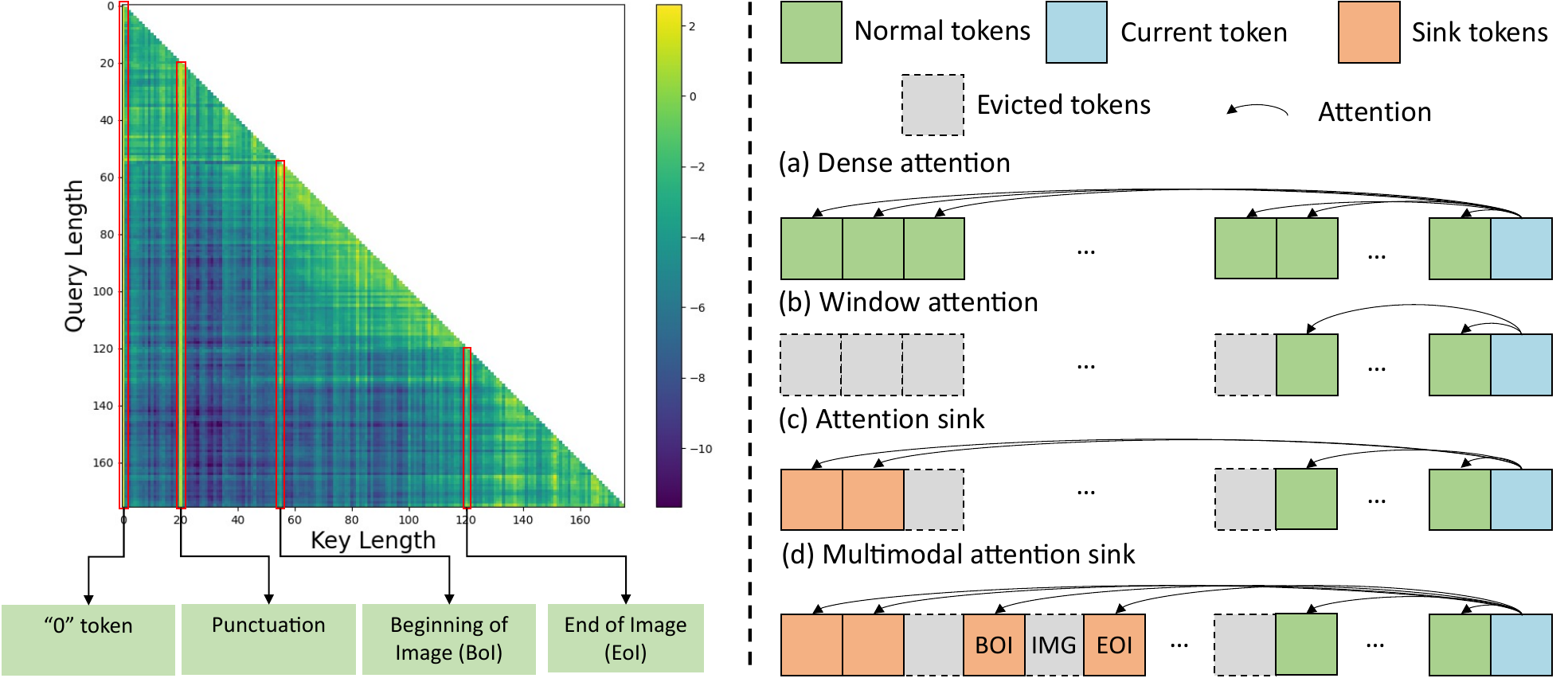}
\caption{\textbf{Left:} Visualization of the attention map when predicting the next token for multimodal story generation. We observe that important attentions are aggregated into the first token of the whole sequence (``0'' token), punctuation tokens, tokens adjacent to BoI, and tokens adjacent to EoI. \textbf{Right:} The diagram of (a) Dense attention, which preserves all tokens in KV cache. (b) Window attention, which evicts preceding tokens by a sliding window. (c) Attention sink, which preserves the beginning tokens based on window attention. (d) Multimodal attention sink, which preserves the beginning of text tokens, images tokens, and the end of image token based on window attention. It can efficiently enable our model to generalize to generating longer sequences than the training sequence length.}
%\caption{\textbf{Left:} Visualization of the attention map when generating next story image and text based on inputs. Important attentions are aggregated into the first token of the whole sequence (``0'' token), punctuation tokens, tokens adjacent to BoI, and tokens adjacent to EoI. \textbf{Right:} (a) Dense attention. Preserving all tokens in KV cache. (b) Window attention. Evicting beginning tokens by a sliding window. (c) Attention sink. Preserving the beginning tokens and evicts middle tokens. It is proposed in StreamingLLM for efficient long text generation. (d) Multimodal attention sink. Preserving the beginning tokens and the beginning and end of image tokens. It aims to efficiently generate long story image and text.}
\label{fig:img_attn_sink}
\end{figure}

%\vspace{-0.1in}
\paragraph{De-tokenizer Adaptation}
After instruction tuning, the SEED-Story MLLM effectively produces story images with correct semantics but lacks style consistency and details. We attribute this issue to the misalignment between the latent space of the MLLM output and the image features. To address this, we perform de-tokenizer adaptation for style and texture alignment. In this stage, only the SD-XL image de-tokenizer is trained. Conditioned on the MLLM output embeddings, SD-XL is expected to generate images that are pixel-level aligned with the ground truth. The separate training of the de-tokenizer offers two key advantages. First, it avoids optimization conflicts between the LLM and the de-tokenizer. Second, it conserves memory, making the process executable on GPUs with limited memory. Please see more analysis in Section~\ref{sec:ablation_detokenizer_adaptation} of our appendix.

%\vspace{-0.1in}
% inference
\subsection{Long Story Generation with Multimodal Attention Sink}
% motivation
Generating long visual stories has substantial potential in various applications, including education and entertainment. However, creating these stories with MLLMs presents significant challenges. Datasets for extended, interleaved stories are not only rare but also impede the training process due to their complexity. To address this, we have to employ a train-short-test-long approach, training models on shorter narratives and extending to longer generations during inference.

Moreover, during inference, generating significantly longer stories than the training data often leads to model degradation, producing lower-quality images, as illustrated in the first row of Figure~\ref{fig:attn_sink_exp}. This process also requires extensive token usage to ensure continuity and coherence, which in turn increases memory and computational demands.

A simplistic solution for this is to use a sliding window technique, depicted in Figure~\ref{fig:img_attn_sink} right (b). However, this method disrupts the token relationships in the Key-Value (KV) cache, resulting in subpar generative outcomes, as demonstrated by StreamingLLM~\cite{xiao2023efficient}. To overcome this, StreamingLLM introduces an attention sink mechanism that preserves the initial tokens, thus allowing for efficient processing of lengthy generations without quality compromise. While effective in language models, its efficacy diminishes in multimodal contexts, as shown in Figure~\ref{fig:img_attn_sink} right (c).

To enhance long multimodal generation, we revisit the attention maps of MLLMs. After conducting numerous experiments (see Section~\ref{sec:appendix_attn_sink} of the appendix for more details) across various models and cases, we analyze the attention maps across different layers and heads. Our analysis reveals that most queries predominantly focus on four types of tokens: (1) starting tokens, (2) punctuation tokens, (3) beginning-of-image (BoI) tokens, and (4) end-of-image (EoI) tokens. Unlike language-only models, MLLMs place considerable attention on specific image tokens, particularly those near the BoI and EoI, as illustrated in Figure~\ref{fig:img_attn_sink} left.

Building on these insights, we propose a new mechanism for extended generation in MLLMs, termed the multimodal attention sink. During generation, we consistently retain the starting tokens and the image tokens adjacent to the BoI and EoI. Although punctuation tokens receive high attention values, their latent value norms are minimal, contributing insignificantly to the final output, so we do not keep them, as noted by \cite{ge2023model}. Our proposed mechanism enables our model to generate high-quality images while maintaining a low computational footprint.

% However, creating long visual stories with generative AI models are extremely costy, since it need the model to remember the information of previous images. For example, ARLDM save previous text and images into BLIP embeddings to instruction following generation. StoryGen use Diffusion intermediate noisy latent to conduct cross attention for latter generation. MLLMs, on the other hand, use token to preserve these information. When story length become longer and longer, the information need to be preserved increase, consuming lots of memory and computational time.

% problem: OOM, low speed

% existing ways: Slide window, attention sink

% We revisit attn map and find xxx

% Why sink to image, important

% we design img attn sink
\begin{figure}[h!]
\centering
\includegraphics[width=1.\linewidth]{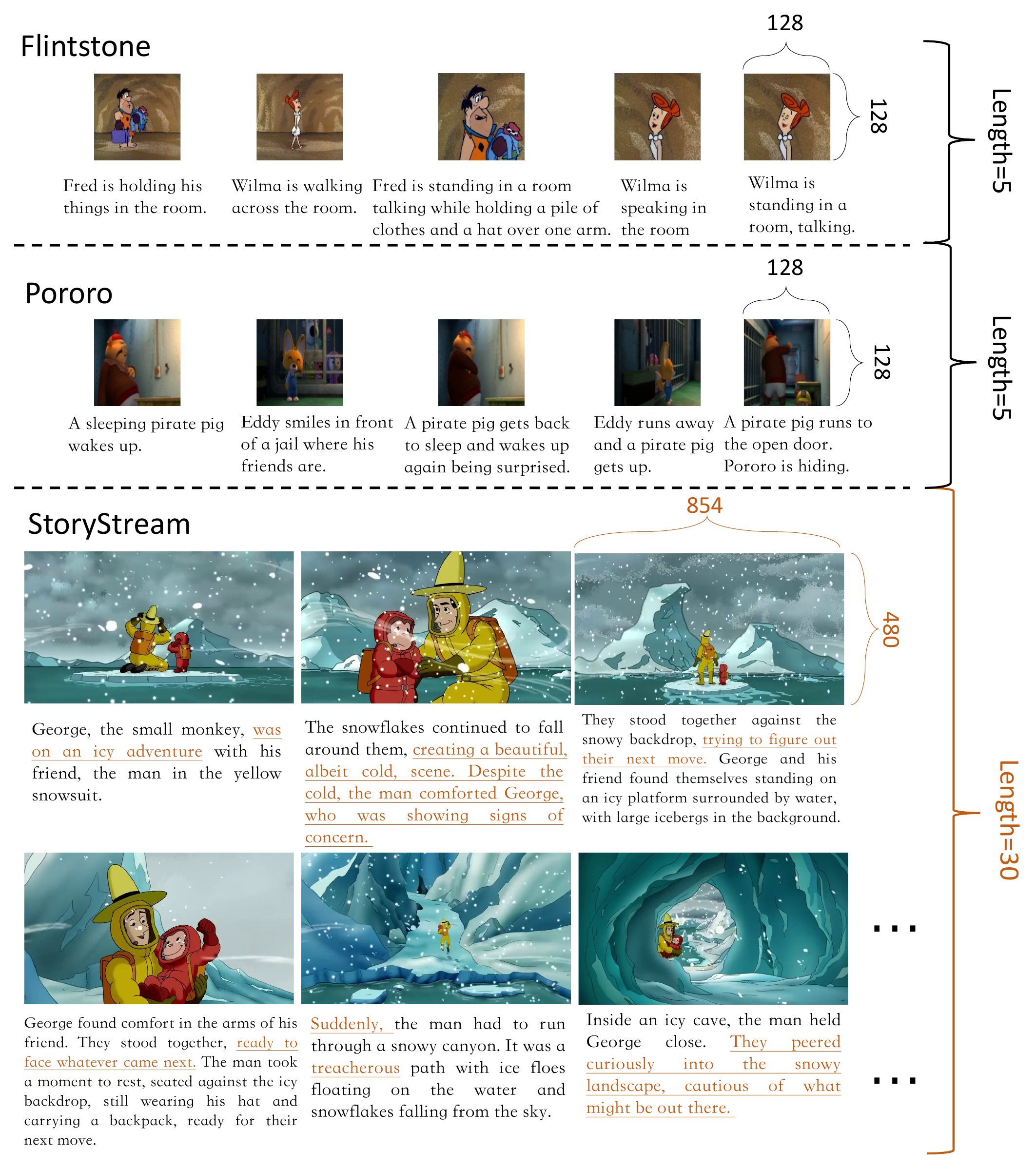}
\caption{Data sample of our StoryStream dataset and existing multimodal story generation datasets. Our multimodal story sequences consist of high-resolution images that are visually engaging, and detailed narrative texts as underlined, closely resembling the real-world storybooks. Additionally, our stories are more extended in length.}
%The data sample from our StoryStream dataset, compared to existing multimodal story generation datasets, demonstrates several distinct enhancements. Our multimodal sequences comprise high-resolution images that are visually engaging, accompanied by detailed narrative texts, as underlined, closely mirroring the storytelling style found in real-world storybooks. Additionally, our stories are more extended in length, providing a richer and more immersive storytelling experience.
% \caption{Our StoryStream dataset is visualized alongside existing datasets. Our images are highlighted with high resolution and detailed granularity. Our text descriptions are vivid and engaging, as underlined, resembling the storytelling found in real storybooks. Our story length and plot are more extended.}
\label{fig:dataset_vis}
%\vspace{-0.3in}
\end{figure}

\begin{figure}[h]
\centering
\includegraphics[width=1.0\linewidth]{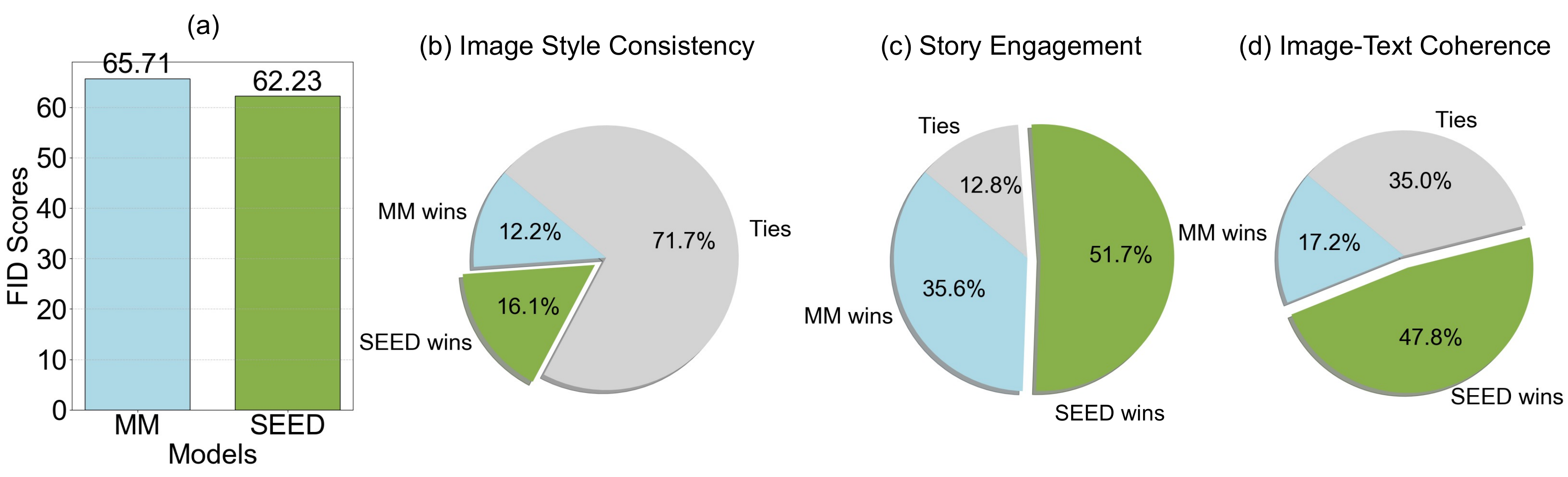}
\caption{Quantitative evaluation of multimodal story generation between MM-interleaved versus SEED-Story. \textbf{(a):} Histograms displaying FID scores. \textbf{(b, c, d):} GPT-4V is employed to choose a preferred result generated by MM-interleaved and SEED-Story respectively. Pie charts show the win rate, where ``Ties'' indicates GPT-4V assesses their outcomes with equal scores.}
% \caption{Quantitative evaluation of MM-interleaved versus SEED-Story. \textbf{(a)}: Histograms depicting FID scores. \textbf{(b, c, d):} Evaluation of MM-interleaved versus SEED-Story. Pie charts displays the results of a comparative analysis where GPT was used to assess the quality of stories. The evaluation focuses on three key aspects: Style Consistency, Story Engagement, and Image-Text Coherence. Win rates for each category show which method GPT frequently favored in generating more effective stories.}
\label{fig:story_gen_quant}
\end{figure}
%\vspace{-0.2in}

\vspace{-0.1in}
\section{StoryStream Dataset}
\vspace{-0.1in}
\subsection{Dataset Construction}
\vspace{-0.1in}
An ideal source for creating a multimodal story generation dataset is cartoon series, which inherently contain rich plots and consistent character portrayals. We selected three cartoon series to construct our dataset and we present the Curious George in the main body of our paper. The process begins with collecting various series, from which we extract keyframes and their associated subtitles~\cite{kilian-2023-video2dataset}. Each keyframe is then processed by GPT-4V~\cite{openai2023gpt4v} or Qwen-VL~\cite{bai2023qwenvl} to generate a detailed image description. These elements—keyframe, subtitle, and description—are compiled into a single group. We aggregate 30 such groups and input them into GPT-4, supplemented with background information about the cartoon series. Following our instructions, GPT-4 generates high-quality narrative texts suitable for training story generation models.

During dataset construction, we discovered that employing the above chain of thought approach not only produces more accurate narrative text but also speeds up the construction process. Unlike directly feeding all images directly to GPT-4, which is limited to 10 images due to API constrains, our approach produces longer stories. We also significantly improve the model’s understanding of each image by incorporating detailed descriptions. This enhancement in image comprehension enriches the narrative details, providing a richer story generation reference. 

% \begin{wrapfigure}{r}{0.35\textwidth} % This aligns the figure to the right and sets it to half the text width
%   \centering
%   \includegraphics[width=\linewidth]{figs/dataset_construction.pdf}
%   \caption{Pipeline of constructing StoryStream dataset.}
%   \label{fig:dataset_construction}
% \end{wrapfigure}
% \vspace{-0.1in}

\subsection{Key Features}

\paragraph{Large-scale.}
Our StoryStream dataset comprises three subsets totaling 257,850 released images. This represents a significant improvement over existing datasets in terms of scale, specific numbers are presented in column 2 of Table~\ref{tab:visual_story_datasets}. To the best of my knowledge, it is the largest visual story generation dataset featuring consistent main characters.

%\vspace{-0.15in}
\paragraph{High Resolution.}
Unlike existing story generation datasets which offer images at a resolution of 128x128, our story stream dataset provides high-resolution images of 480x768.
%\vspace{-0.15in}

\paragraph{Narrative Text.}
Our dataset diverges from existing ones that utilize simple and descriptive language. We offer abstract, narrative, detailed, and story-toned texts that are more akin to real-world applications, such as visualizing narratives from a storybook, examples are shown in Figure~\ref{fig:dataset_vis}. Story text of existing datasets obey the form of ``name'' + ``action'', like ``Poby is playing the violin.''. Contrarily, our story text involves more intrinsic elements.
This effectively enhances the engagement of audiences. An analysis of the average text length per sentence is shown in column 5 of Table~\ref{tab:visual_story_datasets}.
%\vspace{-0.15in}

\paragraph{Long Sequence.}
Moreover, our dataset enhances long story comprehension by offering up to 30 images per story point. Within these 30 images, our corresponding texts present a cohesive narrative, effectively conveying the progression and intricacies of extended stories.
%\vspace{-0.15in}

\begin{figure}[h]
\centering
\includegraphics[width=.9\linewidth]{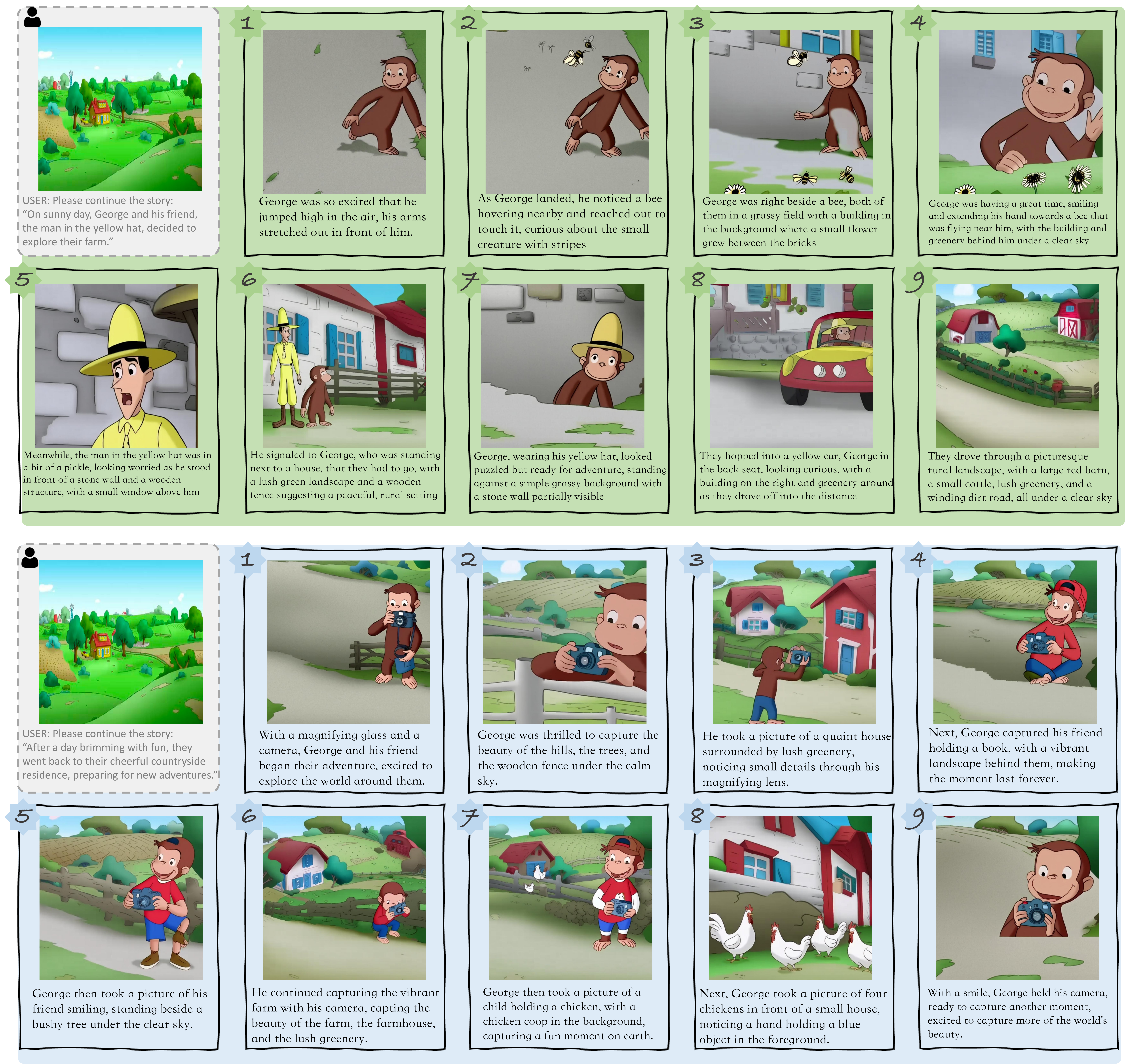}
\caption{Examples of multimodal story generation from SEED-Story with the same initial image but different texts as the beginning. The top branch starts with the text referencing ``the man in the yellow hat'', which leads to the generation of images that include the character. The bottom branch starts without mentioning the man, resulting in a different storyline.}
\label{fig:multimodal_story_gen_control}
\end{figure}

\section{Experiment}

% MM interleaved
\subsection{Quantitative Results}
For quantitative evaluation of multimodal story generation, since there are relatively few established methods for generating multimodal stories, we first fine-tune the recently developed MM-interleaved model on our training dataset for a fair comparison as the baseline model. The quantitative results are listed in Figure~\ref{fig:story_gen_quant}.
%For comparative analysis, there are relatively few established methods for generating multimodal stories. To establish a baseline for comparison, we fine-tune the recently developed MM-interleaved model on our dataset. We detail the comparative results in Figure~\ref{fig:story_gen_quant}. 
The FID is employed to assess the visual quality of the generated images. Additionally, we ask GPT-4V (``gpt-4-turbo-2024-04-09'') to compare and choose a preferred option between each of the generation results of MM-interleaved and SEED-Story across several dimensions: (a) Style Consistency, which evaluates the stylistic uniformity across different images; (b) Story Engagement, which measures the ability of narratives to captivate and maintain audience interest; (c) Image-Text Coherence, which assesses the alignment and relevance between images and their accompanying texts. The details of evaluation are introduced in Section~\ref{sec:evaluation} of Appendix. 
We also conduct a user study, which compares SEED-Story and MM-Interleaved in Section~\ref{sec:user_study} of Appendix. 
The quantitative evaluations demonstrate that SEED-Story outperforms the baseline model in terms of story engagement and image-text coherence, and achieves slightly higher preference in image style consistency. We also provide quantitative comparisons of story
visualization in Section~\ref{sec:story_visualization} of Appendix.

\begin{figure}[h]
\centering
% \vspace{-0.1in}
\includegraphics[width=1.0\linewidth]{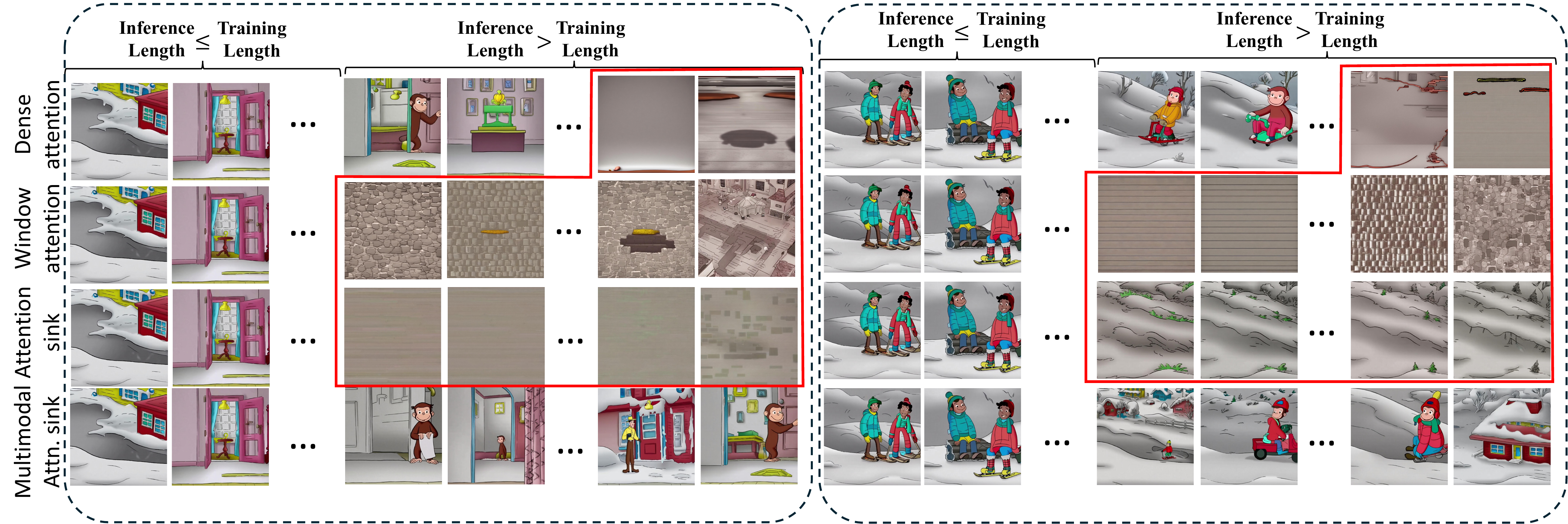}
\caption{The visualization of generating long stories with different attention mechanisms. Without multimodal attention sink, MLLM cannot generate meaningful long image sequences. As highlighted in the red boxes, other methods produce meaningless images in the later frames.}
% \caption{The visualization of generating long stories with different attention mechanisms. Other mechanisms collapse when producing long sequence, while multimodal attention sink consistently generates high-quality images.}
\label{fig:attn_sink_exp}
\end{figure}
\vspace{-0.3in}

\begin{table}[h]
\centering
\caption{Quantitative evaluation of long story generation with various attention mechanisms. FID and CLIP scores are calculated by comparing ground truth images with generated images. Inference time and memory usage are calculated by generating 50 sequences multiple times for average.}
% \caption{Quantitative evaluation of various attention mechanisms in generating long visual stories. FID and CLIP scores are calculated by comparing ground truth images with generated images. Inference time and memory usage are assessed through multiple time averages taken during the generation of 50-image stories.}
\begin{tabular}{lcccc}
\hline
\textbf{Metrics} & \textbf{FID $\downarrow$} & \textbf{CLIP Score $\uparrow$} & \textbf{Inference Time (s) $\downarrow$} & \textbf{Memory (GB) $\downarrow$} \\
\hline
\textbf{Dense Attn} & 119.72 & 0.705 & 569.67 & 37.99 \\
\textbf{Window Attn} & 334.90 & 0.598 & \textbf{450.64} & \textbf{30.81} \\
\textbf{Attn Sink} & 221.53 & 0.676 & 451.94 & \textbf{30.81} \\
\textbf{Multimodal Attn Sink} & \textbf{79.67} & \textbf{0.728} & 473.98 & 31.82 \\
\hline
\end{tabular}

\vspace{-0.2in}
\label{tab:attn_sink_exp}
\end{table}

\subsection{Qualitative Results}
% \vspace{-0.1in}
For qualitative evaluations, we first demonstrate that our SEED-story can effectively generate stories with different plots and corresponding illustrations based on the different beginnings provided by the user, as shown in Fig.~\ref{fig:multimodal_story_gen_control}. We provide more qualitative results of multimodal story generation in Section~\ref{sec:visual_results} of Appendix. As shown in Figure~\ref{fig:multimodal_story_gen_1}, Figure~\ref{fig:multimodal_story_gen_2} and Figure~\ref{fig:sup_gen_case3}, SEED-story can generate long sequences with engaging plots and vivid images.  We further provide qualitative comparisons of story
visualization in Section~\ref{sec:story_visualization} of Appendix.

\vspace{-0.1in}
\subsection{Multimodal Attention Sink}
% other methods, dense attn, window attn, streamingLLM, ours
% metrics
To verify the effectiveness of multimodal attention sink in long story generation, we conduct an experiment visualizing a long story using the SEED-Story model, but with varying attention mechanisms. We chunk our data into stories of length of 10 considering the training efficiency. We set the window size as the same as the training length. Qualitative results presented in Figure~\ref{fig:attn_sink_exp} demonstrate that window attention quickly collapses when the inference length exceeds the training length. Both dense attention and attention sink approaches fare better, yet still fail to produce meaningful images as the inference sequence lengthens. In contrast, the multimodal attention sink consistently produces high-quality images. In terms of efficiency, the multimodal attention sink exhibits significant improvement over dense attention, with only a modest increase in time and memory costs compared to window attention and vanilla attention sink. These additional costs stem from retaining extra image tokens in the KV cache. Quantitative results presented in Table~\ref{tab:attn_sink_exp} substantiate the above conclusion. 
\vspace{-0.1in}
\section{Conclusion}
This work introduces SEED-Story, a pioneering approach that leverages a Multimodal Large Language Model to generate multimodal long stories with rich narrative text and contextually relevant images. We propose a multimodal attention sink mechanism to enable our model to generalize to generating long sequences in an efficient manner. We further present a high-quality dataset named StoryStream for training and benchmarking the task of multimodal story generation effectively.

\bibliographystyle{iclr2025_conference}
\bibliography{iclr2025_conference}

\begin{thebibliography}{64}
\providecommand{\natexlab}[1]{#1}
\providecommand{\url}[1]{\texttt{#1}}
\expandafter\ifx\csname urlstyle\endcsname\relax
  \providecommand{\doi}[1]{doi: #1}\else
  \providecommand{\doi}{doi: \begingroup \urlstyle{rm}\Url}\fi

\bibitem[Aiello et~al.(2023)Aiello, Yu, Nie, Aghajanyan, and Oguz]{aiello2023jointly}
Emanuele Aiello, Lili Yu, Yixin Nie, Armen Aghajanyan, and Barlas Oguz.
\newblock Jointly training large autoregressive multimodal models.
\newblock \emph{arXiv preprint arXiv:2309.15564}, 2023.

\bibitem[An et~al.(2023)An, Gong, Zhong, Li, Zhang, Kong, and Qiu]{an2023leval}
Chenxin An, Shansan Gong, Ming Zhong, Mukai Li, Jun Zhang, Lingpeng Kong, and Xipeng Qiu.
\newblock L-eval: Instituting standardized evaluation for long context language models, 2023.

\bibitem[{Animaj}(2024{\natexlab{a}})]{Animaj}
{Animaj}.
\newblock Animaj official website, 2024{\natexlab{a}}.
\newblock URL \url{https://www.animaj.com/#pocoyo/}.
\newblock Accessed: 2024-05-22.

\bibitem[{Animaj}(2024{\natexlab{b}})]{RabbidsInvasionYouTube}
{Animaj}.
\newblock Rabbids invasion official youtube channel, 2024{\natexlab{b}}.
\newblock URL \url{https://www.youtube.com/@RabbidsInvasion}.
\newblock Accessed: 2024-05-22.

\bibitem[Avrahami et~al.(2024)Avrahami, Hertz, Vinker, Arar, Fruchter, Fried, Cohen-Or, and Lischinski]{avrahami2024chosen}
Omri Avrahami, Amir Hertz, Yael Vinker, Moab Arar, Shlomi Fruchter, Ohad Fried, Daniel Cohen-Or, and Dani Lischinski.
\newblock The chosen one: Consistent characters in text-to-image diffusion models.
\newblock In \emph{ACM SIGGRAPH 2024 Conference Papers}, pp.\  1--12, 2024.

\bibitem[Bai et~al.(2023{\natexlab{a}})Bai, Bai, Yang, Wang, Tan, Wang, Lin, Zhou, and Zhou]{bai2023qwen}
Jinze Bai, Shuai Bai, Shusheng Yang, Shijie Wang, Sinan Tan, Peng Wang, Junyang Lin, Chang Zhou, and Jingren Zhou.
\newblock Qwen-vl: A versatile vision-language model for understanding, localization, text reading, and beyond.
\newblock 2023{\natexlab{a}}.

\bibitem[Bai et~al.(2023{\natexlab{b}})Bai, Bai, Yang, Wang, Tan, Wang, Lin, Zhou, and Zhou]{bai2023qwenvl}
Jinze Bai, Shuai Bai, Shusheng Yang, Shijie Wang, Sinan Tan, Peng Wang, Junyang Lin, Chang Zhou, and Jingren Zhou.
\newblock Qwen-vl: A versatile vision-language model for understanding, localization, text reading, and beyond, 2023{\natexlab{b}}.

\bibitem[Beltagy et~al.(2020)Beltagy, Peters, and Cohan]{beltagy2020longformer}
Iz~Beltagy, Matthew~E Peters, and Arman Cohan.
\newblock Longformer: The long-document transformer.
\newblock \emph{arXiv preprint arXiv:2004.05150}, 2020.

\bibitem[Brown et~al.(2020)Brown, Mann, Ryder, Subbiah, Kaplan, Dhariwal, Neelakantan, Shyam, Sastry, Askell, et~al.]{brown2020language}
Tom Brown, Benjamin Mann, Nick Ryder, Melanie Subbiah, Jared~D Kaplan, Prafulla Dhariwal, Arvind Neelakantan, Pranav Shyam, Girish Sastry, Amanda Askell, et~al.
\newblock Language models are few-shot learners.
\newblock In \emph{Advances in Neural Information Processing Systems}, volume~33, pp.\  1877--1901, 2020.

\bibitem[Chen et~al.(2023)Chen, Yu, Ge, Yao, Xie, Wu, Wang, Kwok, Luo, Lu, and Li]{chen2023pixartalpha}
Junsong Chen, Jincheng Yu, Chongjian Ge, Lewei Yao, Enze Xie, Yue Wu, Zhongdao Wang, James Kwok, Ping Luo, Huchuan Lu, and Zhenguo Li.
\newblock Pixart-$\alpha$: Fast training of diffusion transformer for photorealistic text-to-image synthesis, 2023.

\bibitem[Chowdhery et~al.(2022)Chowdhery, Narang, Devlin, Bosma, Mishra, Roberts, Barham, Chung, Sutton, Gehrmann, et~al.]{chowdhery2022palm}
Aakanksha Chowdhery, Sharan Narang, Jacob Devlin, Maarten Bosma, Gaurav Mishra, Adam Roberts, Paul Barham, Hyung~Won Chung, Charles Sutton, Sebastian Gehrmann, et~al.
\newblock Palm: Scaling language modeling with pathways.
\newblock \emph{arXiv preprint arXiv:2204.02311}, 2022.

\bibitem[Dong et~al.(2023)Dong, Han, Peng, Qi, Ge, Yang, Zhao, Sun, Zhou, Wei, et~al.]{dong2023dreamllm}
Runpei Dong, Chunrui Han, Yuang Peng, Zekun Qi, Zheng Ge, Jinrong Yang, Liang Zhao, Jianjian Sun, Hongyu Zhou, Haoran Wei, et~al.
\newblock Dreamllm: Synergistic multimodal comprehension and creation.
\newblock \emph{arXiv preprint arXiv:2309.11499}, 2023.

\bibitem[Dosovitskiy et~al.(2020)Dosovitskiy, Beyer, Kolesnikov, Weissenborn, Zhai, Unterthiner, Dehghani, Minderer, Heigold, Gelly, et~al.]{dosovitskiy2020image}
Alexey Dosovitskiy, Lucas Beyer, Alexander Kolesnikov, Dirk Weissenborn, Xiaohua Zhai, Thomas Unterthiner, Mostafa Dehghani, Matthias Minderer, Georg Heigold, Sylvain Gelly, et~al.
\newblock An image is worth 16x16 words: Transformers for image recognition at scale.
\newblock \emph{arXiv preprint arXiv:2010.11929}, 2020.

\bibitem[Ge et~al.(2023{\natexlab{a}})Ge, Zhang, Liu, Zhang, Han, and Gao]{ge2023model}
Suyu Ge, Yunan Zhang, Liyuan Liu, Minjia Zhang, Jiawei Han, and Jianfeng Gao.
\newblock Model tells you what to discard: Adaptive kv cache compression for llms.
\newblock \emph{arXiv preprint arXiv:2310.01801}, 2023{\natexlab{a}}.

\bibitem[Ge et~al.(2023{\natexlab{b}})Ge, Ge, Zeng, Wang, and Shan]{ge2023planting}
Yuying Ge, Yixiao Ge, Ziyun Zeng, Xintao Wang, and Ying Shan.
\newblock Planting a seed of vision in large language model.
\newblock \emph{arXiv preprint arXiv:2307.08041}, 2023{\natexlab{b}}.

\bibitem[Ge et~al.(2023{\natexlab{c}})Ge, Zhao, Zeng, Ge, Li, Wang, and Shan]{ge2023making}
Yuying Ge, Sijie Zhao, Ziyun Zeng, Yixiao Ge, Chen Li, Xintao Wang, and Ying Shan.
\newblock Making llama see and draw with seed tokenizer, 2023{\natexlab{c}}.

\bibitem[Ge et~al.(2024)Ge, Zhao, Zhu, Ge, Yi, Song, Li, Ding, and Shan]{ge2024seedx}
Yuying Ge, Sijie Zhao, Jinguo Zhu, Yixiao Ge, Kun Yi, Lin Song, Chen Li, Xiaohan Ding, and Ying Shan.
\newblock Seed-x: Multimodal models with unified multi-granularity comprehension and generation, 2024.

\bibitem[Gong et~al.(2023)Gong, Pang, Cun, Xia, Chen, Wang, Zhang, Wang, Shan, and Yang]{gong2023talecrafter}
Yuan Gong, Youxin Pang, Xiaodong Cun, Menghan Xia, Haoxin Chen, Longyue Wang, Yong Zhang, Xintao Wang, Ying Shan, and Yujiu Yang.
\newblock Talecrafter: Interactive story visualization with multiple characters.
\newblock \emph{arXiv preprint arXiv:2305.18247}, 2023.

\bibitem[Gupta et~al.(2018)Gupta, Schwenk, Farhadi, Hoiem, and Kembhavi]{gupta2018imagine}
Tanmay Gupta, Dustin Schwenk, Ali Farhadi, Derek Hoiem, and Aniruddha Kembhavi.
\newblock Imagine this! scripts to compositions to videos.
\newblock In \emph{Proceedings of the European conference on computer vision (ECCV)}, pp.\  598--613, 2018.

\bibitem[Hu et~al.(2021)Hu, Shen, Wallis, Allen-Zhu, Li, Wang, Wang, and Chen]{hu2021lora}
Edward~J Hu, Yelong Shen, Phillip Wallis, Zeyuan Allen-Zhu, Yuanzhi Li, Shean Wang, Lu~Wang, and Weizhu Chen.
\newblock Lora: Low-rank adaptation of large language models.
\newblock \emph{arXiv preprint arXiv:2106.09685}, 2021.

\bibitem[Huang et~al.(2016)Huang, Ferraro, Mostafazadeh, Misra, Agrawal, Devlin, Girshick, He, Kohli, Batra, et~al.]{huang2016visual}
Ting-Hao Huang, Francis Ferraro, Nasrin Mostafazadeh, Ishan Misra, Aishwarya Agrawal, Jacob Devlin, Ross Girshick, Xiaodong He, Pushmeet Kohli, Dhruv Batra, et~al.
\newblock Visual storytelling.
\newblock In \emph{Proceedings of the 2016 conference of the North American chapter of the association for computational linguistics: Human language technologies}, pp.\  1233--1239, 2016.

\bibitem[Kilian et~al.(2023)Kilian, Beaumont, Mendelevitch, Kulal, and Blattmann]{kilian-2023-video2dataset}
Maciej Kilian, Romain Beaumont, Daniel Mendelevitch, Sumith Kulal, and Andreas Blattmann.
\newblock video2dataset: Easily turn large sets of video urls to a video dataset.
\newblock \url{https://github.com/iejMac/video2dataset}, 2023.

\bibitem[Lauren{\c{c}}on et~al.(2024)Lauren{\c{c}}on, Saulnier, Tronchon, Bekman, Singh, Lozhkov, Wang, Karamcheti, Rush, Kiela, et~al.]{laurenccon2024obelics}
Hugo Lauren{\c{c}}on, Lucile Saulnier, L{\'e}o Tronchon, Stas Bekman, Amanpreet Singh, Anton Lozhkov, Thomas Wang, Siddharth Karamcheti, Alexander Rush, Douwe Kiela, et~al.
\newblock Obelics: An open web-scale filtered dataset of interleaved image-text documents.
\newblock \emph{Advances in Neural Information Processing Systems}, 36, 2024.

\bibitem[Li et~al.(2023)Li, Li, Savarese, and Hoi]{li2023blip2}
Junnan Li, Dongxu Li, Silvio Savarese, and Steven Hoi.
\newblock Blip-2: Bootstrapping language-image pre-training with frozen image encoders and large language models.
\newblock In \emph{Proceedings of the International Conference on Machine Learning (ICML)}, 2023.

\bibitem[Li et~al.(2024)Li, Zhang, Wang, Zhong, Chen, Chu, Liu, and Jia]{li2024minigemini}
Yanwei Li, Yuechen Zhang, Chengyao Wang, Zhisheng Zhong, Yixin Chen, Ruihang Chu, Shaoteng Liu, and Jiaya Jia.
\newblock Mini-gemini: Mining the potential of multi-modality vision language models.
\newblock \emph{arXiv preprint arXiv:2403.18814}, 2024.

\bibitem[Li et~al.(2019)Li, Gan, Shen, Liu, Cheng, Wu, Carin, Carlson, and Gao]{li2019storygan}
Yitong Li, Zhe Gan, Yelong Shen, Jingjing Liu, Yu~Cheng, Yuexin Wu, Lawrence Carin, David Carlson, and Jianfeng Gao.
\newblock Storygan: A sequential conditional gan for story visualization.
\newblock In \emph{Proceedings of the IEEE/CVF Conference on Computer Vision and Pattern Recognition}, pp.\  6329--6338, 2019.

\bibitem[Lin et~al.(2023)Lin, Liu, Zhang, Gao, Qiu, Xiao, Qiu, Lin, Shao, Chen, et~al.]{lin2023sphinx}
Ziyi Lin, Chris Liu, Renrui Zhang, Peng Gao, Longtian Qiu, Han Xiao, Han Qiu, Chen Lin, Wenqi Shao, Keqin Chen, et~al.
\newblock Sphinx: The joint mixing of weights, tasks, and visual embeddings for multi-modal large language models.
\newblock \emph{arXiv preprint arXiv:2311.07575}, 2023.

\bibitem[Liu et~al.(2023{\natexlab{a}})Liu, Wu, Zhong, Zhang, and Xie]{liu2023intelligent}
Chang Liu, Haoning Wu, Yujie Zhong, Xiaoyun Zhang, and Weidi Xie.
\newblock Intelligent grimm--open-ended visual storytelling via latent diffusion models.
\newblock \emph{arXiv preprint arXiv:2306.00973}, 2023{\natexlab{a}}.

\bibitem[Liu et~al.(2023{\natexlab{b}})Liu, Li, Li, and Lee]{liu2023improved}
Haotian Liu, Chunyuan Li, Yuheng Li, and Yong~Jae Lee.
\newblock Improved baselines with visual instruction tuning.
\newblock \emph{arXiv preprint arXiv:2310.03744}, 2023{\natexlab{b}}.

\bibitem[Liu et~al.(2023{\natexlab{c}})Liu, Li, Li, and Lee]{liu2023llava1.5}
Haotian Liu, Chunyuan Li, Yuheng Li, and Yong~Jae Lee.
\newblock Improved baselines with visual instruction tuning.
\newblock \emph{arXiv preprint arXiv:2310.03744}, 2023{\natexlab{c}}.

\bibitem[Liu et~al.(2023{\natexlab{d}})Liu, Li, Wu, and Lee]{liu2023visual}
Haotian Liu, Chunyuan Li, Qingyang Wu, and Yong~Jae Lee.
\newblock Visual instruction tuning.
\newblock \emph{arXiv preprint arXiv:2304.08485}, 2023{\natexlab{d}}.

\bibitem[Maharana \& Bansal(2021)Maharana and Bansal]{maharana2021integrating}
Adyasha Maharana and Mohit Bansal.
\newblock Integrating visuospatial, linguistic and commonsense structure into story visualization.
\newblock \emph{arXiv preprint arXiv:2110.10834}, 2021.

\bibitem[Maharana et~al.(2021)Maharana, Hannan, and Bansal]{maharana2021improving}
Adyasha Maharana, Darryl Hannan, and Mohit Bansal.
\newblock Improving generation and evaluation of visual stories via semantic consistency.
\newblock \emph{arXiv preprint arXiv:2105.10026}, 2021.

\bibitem[Maharana et~al.(2022)Maharana, Hannan, and Bansal]{maharana2022storydall}
Adyasha Maharana, Darryl Hannan, and Mohit Bansal.
\newblock Storydall-e: Adapting pretrained text-to-image transformers for story continuation.
\newblock In \emph{European Conference on Computer Vision}, pp.\  70--87. Springer, 2022.

\bibitem[{OpenAI}(2023)]{openai2023gpt4v}
{OpenAI}.
\newblock Gpt-4v: Optimizing language models for dialogue.
\newblock \url{https://www.openai.com/chatgpt}, 2023.

\bibitem[Pan et~al.(2024)Pan, Qin, Li, Xue, and Chen]{pan2024synthesizing}
Xichen Pan, Pengda Qin, Yuhong Li, Hui Xue, and Wenhu Chen.
\newblock Synthesizing coherent story with auto-regressive latent diffusion models.
\newblock In \emph{Proceedings of the IEEE/CVF Winter Conference on Applications of Computer Vision}, pp.\  2920--2930, 2024.

\bibitem[Patashnik et~al.(2021)Patashnik, Wu, Shechtman, Cohen-Or, and Lischinski]{Patashnik_2021_ICCV}
Or~Patashnik, Zongze Wu, Eli Shechtman, Daniel Cohen-Or, and Dani Lischinski.
\newblock Styleclip: Text-driven manipulation of stylegan imagery.
\newblock In \emph{Proceedings of the IEEE/CVF International Conference on Computer Vision (ICCV)}, pp.\  2085--2094, October 2021.

\bibitem[{PBS Kids}(2024{\natexlab{a}})]{CuriousGeorgePBSKids}
{PBS Kids}.
\newblock Curious george official website, 2024{\natexlab{a}}.
\newblock URL \url{https://pbskids.org/curiousgeorge/}.
\newblock Accessed: 2024-05-22.

\bibitem[{PBS Kids}(2024{\natexlab{b}})]{CuriousGeorgeYouTube}
{PBS Kids}.
\newblock Curious george official youtube channel, 2024{\natexlab{b}}.
\newblock URL \url{https://www.youtube.com/@CuriousGeorge}.
\newblock Accessed: 2024-05-22.

\bibitem[Peng et~al.(2023{\natexlab{a}})Peng, Wang, Dong, Hao, Huang, Ma, and Wei]{peng2023kosmos}
Zhiliang Peng, Wenhui Wang, Li~Dong, Yaru Hao, Shaohan Huang, Shuming Ma, and Furu Wei.
\newblock Kosmos-2: Grounding multimodal large language models to the world.
\newblock \emph{arXiv preprint arXiv:2306.14824}, 2023{\natexlab{a}}.

\bibitem[Peng et~al.(2023{\natexlab{b}})Peng, Wang, Dong, Hao, Huang, Ma, and Wei]{peng2023kosmos2}
Zhiliang Peng, Wenhui Wang, Li~Dong, Yaru Hao, Shaohan Huang, Shuming Ma, and Furu Wei.
\newblock Kosmos-2: Grounding multimodal large language models to the world.
\newblock \emph{arXiv preprint arXiv:2306.14824}, 2023{\natexlab{b}}.

\bibitem[Podell et~al.(2023)Podell, English, Lacey, Blattmann, Dockhorn, M{\"u}ller, Penna, and Rombach]{podell2023sdxl}
Dustin Podell, Zion English, Kyle Lacey, Andreas Blattmann, Tim Dockhorn, Jonas M{\"u}ller, Joe Penna, and Robin Rombach.
\newblock Sdxl: Improving latent diffusion models for high-resolution image synthesis.
\newblock \emph{arXiv preprint arXiv:2307.01952}, 2023.

\bibitem[Rahman et~al.(2023)Rahman, Lee, Ren, Tulyakov, Mahajan, and Sigal]{rahman2023make}
Tanzila Rahman, Hsin-Ying Lee, Jian Ren, Sergey Tulyakov, Shweta Mahajan, and Leonid Sigal.
\newblock Make-a-story: Visual memory conditioned consistent story generation.
\newblock In \emph{Proceedings of the IEEE/CVF Conference on Computer Vision and Pattern Recognition}, pp.\  2493--2502, 2023.

\bibitem[Rombach et~al.(2022)Rombach, Blattmann, Lorenz, Esser, and Ommer]{rombach2022high}
Robin Rombach, Andreas Blattmann, Dominik Lorenz, Patrick Esser, and Bj{\"o}rn Ommer.
\newblock High-resolution image synthesis with latent diffusion models.
\newblock In \emph{Proceedings of the IEEE/CVF conference on computer vision and pattern recognition}, pp.\  10684--10695, 2022.

\bibitem[Ruiz et~al.(2023)Ruiz, Li, Jampani, Pritch, Rubinstein, and Aberman]{ruiz2023dreambooth}
Nataniel Ruiz, Yuanzhen Li, Varun Jampani, Yael Pritch, Michael Rubinstein, and Kfir Aberman.
\newblock Dreambooth: Fine tuning text-to-image diffusion models for subject-driven generation.
\newblock In \emph{Proceedings of the IEEE/CVF conference on computer vision and pattern recognition}, pp.\  22500--22510, 2023.

\bibitem[Shen \& Elhoseiny(2023)Shen and Elhoseiny]{shen2023storygptv}
Xiaoqian Shen and Mohamed Elhoseiny.
\newblock Storygpt-v: Large language models as consistent story visualizers, 2023.

\bibitem[Sun et~al.(2023{\natexlab{a}})Sun, Cui, Zhang, Zhang, Yu, Luo, Wang, Rao, Liu, Huang, and Wang]{Emu2}
Quan Sun, Yufeng Cui, Xiaosong Zhang, Fan Zhang, Qiying Yu, Zhengxiong Luo, Yueze Wang, Yongming Rao, Jingjing Liu, Tiejun Huang, and Xinlong Wang.
\newblock Generative multimodal models are in-context learners.
\newblock 2023{\natexlab{a}}.

\bibitem[Sun et~al.(2023{\natexlab{b}})Sun, Yu, Cui, Zhang, Zhang, Wang, Gao, Liu, Huang, and Wang]{Emu}
Quan Sun, Qiying Yu, Yufeng Cui, Fan Zhang, Xiaosong Zhang, Yueze Wang, Hongcheng Gao, Jingjing Liu, Tiejun Huang, and Xinlong Wang.
\newblock Generative pretraining in multimodality.
\newblock 2023{\natexlab{b}}.

\bibitem[Sun et~al.(2023{\natexlab{c}})Sun, Yu, Cui, Zhang, Zhang, Wang, Gao, Liu, Huang, and Wang]{sun2023generative}
Quan Sun, Qiying Yu, Yufeng Cui, Fan Zhang, Xiaosong Zhang, Yueze Wang, Hongcheng Gao, Jingjing Liu, Tiejun Huang, and Xinlong Wang.
\newblock Generative pretraining in multimodality.
\newblock \emph{arXiv preprint arXiv:2307.05222}, 2023{\natexlab{c}}.

\bibitem[Taori et~al.(2023)Taori, Gulrajani, Zhang, Dubois, Li, Guestrin, Liang, and Hashimoto]{alpaca}
Rohan Taori, Ishaan Gulrajani, Tianyi Zhang, Yann Dubois, Xuechen Li, Carlos Guestrin, Percy Liang, and Tatsunori~B. Hashimoto.
\newblock Stanford alpaca: An instruction-following llama model.
\newblock \url{https://github.com/tatsu-lab/stanford_alpaca}, 2023.

\bibitem[Team(2024)]{team2024chameleon}
Chameleon Team.
\newblock Chameleon: Mixed-modal early-fusion foundation models.
\newblock \emph{arXiv e-prints}, pp.\  arXiv--2405, 2024.

\bibitem[Tewel et~al.(2024)Tewel, Kaduri, Gal, Kasten, Wolf, Chechik, and Atzmon]{tewel2024training}
Yoad Tewel, Omri Kaduri, Rinon Gal, Yoni Kasten, Lior Wolf, Gal Chechik, and Yuval Atzmon.
\newblock Training-free consistent text-to-image generation.
\newblock \emph{ACM Transactions on Graphics (TOG)}, 43\penalty0 (4):\penalty0 1--18, 2024.

\bibitem[{TheLandBeforeTime}(2024{\natexlab{a}})]{TheLandBeforeTime}
{TheLandBeforeTime}.
\newblock The land before time official youtube channel, 2024{\natexlab{a}}.
\newblock URL \url{https://www.youtube.com/@TheLandBeforeTime}.
\newblock Accessed: 2024-05-22.

\bibitem[{TheLandBeforeTime}(2024{\natexlab{b}})]{TheLandBeforeTimeWebsite}
{TheLandBeforeTime}.
\newblock The land before time official website, 2024{\natexlab{b}}.
\newblock URL \url{https://thelandbeforetime.org}.
\newblock Accessed: 2024-05-22.

\bibitem[Tian et~al.(2024)Tian, Zhu, Xiong, Wang, Chen, Wang, Chen, Lu, Lu, Zhou, et~al.]{tian2024mm}
Changyao Tian, Xizhou Zhu, Yuwen Xiong, Weiyun Wang, Zhe Chen, Wenhai Wang, Yuntao Chen, Lewei Lu, Tong Lu, Jie Zhou, et~al.
\newblock Mm-interleaved: Interleaved image-text generative modeling via multi-modal feature synchronizer.
\newblock \emph{arXiv preprint arXiv:2401.10208}, 2024.

\bibitem[Touvron et~al.(2023)Touvron, Lavril, Izacard, Martinet, Lachaux, Lacroix, Rozière, Goyal, Hambro, Azhar, et~al.]{touvron2023llama}
Hugo Touvron, Thibaut Lavril, Gautier Izacard, Xavier Martinet, Marie-Anne Lachaux, Timothée Lacroix, Baptiste Rozière, Naman Goyal, Eric Hambro, Faisal Azhar, et~al.
\newblock Llama: Open and efficient foundation language models.
\newblock \emph{arXiv preprint arXiv:2302.13971}, 2023.

\bibitem[Wang et~al.(2023)Wang, Yang, Liu, and cong Chen]{wang2023steps}
Luozhou Wang, Shuai Yang, Shu Liu, and Ying cong Chen.
\newblock Not all steps are created equal: Selective diffusion distillation for image manipulation, 2023.

\bibitem[Wu et~al.(2023)Wu, Fei, Qu, Ji, and Chua]{wu2023nextgpt}
Shengqiong Wu, Hao Fei, Leigang Qu, Wei Ji, and Tat-Seng Chua.
\newblock Next-gpt: Any-to-any multimodal llm.
\newblock \emph{arXiv preprint arXiv:2309.05519}, 2023.

\bibitem[Xiao et~al.(2023)Xiao, Tian, Chen, Han, and Lewis]{xiao2023efficient}
Guangxuan Xiao, Yuandong Tian, Beidi Chen, Song Han, and Mike Lewis.
\newblock Efficient streaming language models with attention sinks.
\newblock \emph{arXiv preprint arXiv:2309.17453}, 2023.

\bibitem[Yu et~al.(2023)Yu, Shi, Pasunuru, Muller, Golovneva, Wang, Babu, Tang, Karrer, Sheynin, et~al.]{yu2023scaling}
Lili Yu, Bowen Shi, Ramakanth Pasunuru, Benjamin Muller, Olga Golovneva, Tianlu Wang, Arun Babu, Binh Tang, Brian Karrer, Shelly Sheynin, et~al.
\newblock Scaling autoregressive multi-modal models: Pretraining and instruction tuning.
\newblock \emph{arXiv preprint arXiv:2309.02591}, 2023.

\bibitem[Zhang et~al.(2023)Zhang, Dong, Wang, Cao, Xu, Ouyang, Zhao, Ding, Zhang, Duan, Yan, et~al.]{zhang2023internlm}
Pan Zhang, Xiaoyi Dong, Bin Wang, Yuhang Cao, Chao Xu, Linke Ouyang, Zhiyuan Zhao, Shuangrui Ding, Songyang Zhang, Haodong Duan, Hang Yan, et~al.
\newblock Internlm-xcomposer: A vision-language large model for advanced text-image comprehension and composition.
\newblock \emph{arXiv preprint arXiv:2309.15112}, 2023.

\bibitem[Zheng et~al.(2023)Zheng, Chiang, Sheng, Zhuang, Wu, Zhuang, Lin, Li, Li, Xing, Zhang, Gonzalez, and Stoica]{zheng2023judging}
Lianmin Zheng, Wei-Lin Chiang, Ying Sheng, Siyuan Zhuang, Zhanghao Wu, Yonghao Zhuang, Zi~Lin, Zhuohan Li, Dacheng Li, Eric.~P Xing, Hao Zhang, Joseph~E. Gonzalez, and Ion Stoica.
\newblock Judging llm-as-a-judge with mt-bench and chatbot arena, 2023.

\bibitem[Zhu et~al.(2023{\natexlab{a}})Zhu, Chen, Shen, Li, and Elhoseiny]{zhu2023minigpt4}
Deyao Zhu, Jun Chen, Xiaoqian Shen, Xiang Li, and Mohamed Elhoseiny.
\newblock Minigpt-4: Enhancing vision-language understanding with advanced large language models.
\newblock \emph{arXiv preprint arXiv:2304.10592}, 2023{\natexlab{a}}.

\bibitem[Zhu et~al.(2023{\natexlab{b}})Zhu, Ding, Ge, Ge, Zhao, Zhao, Wang, and Shan]{zhu2023vlgpt}
Jinguo Zhu, Xiaohan Ding, Yixiao Ge, Yuying Ge, Sijie Zhao, Hengshuang Zhao, Xiaohua Wang, and Ying Shan.
\newblock Vl-gpt: A generative pre-trained transformer for vision and language understanding and generation.
\newblock \emph{arXiv preprint arXiv:2312.09251}, 2023{\natexlab{b}}.

\end{thebibliography}

\newpage
\section*{Appendix}
\appendix
\setcounter{figure}{0}
\renewcommand{\thefigure}{\Alph{figure}}

\section{User study}
\label{sec:user_study}
\begin{table}[ht]
\centering
\begin{tabular}{l|c|c|c}
\hline
\textbf{Criteria} & \textbf{MM-Interleaved} & \textbf{SEED-Story} & \textbf{Ties} \\ \hline
Image Quality          & 21.3\%  & 66.2\%  & 12.5\%  \\ \hline
Image Style Consistency & 7.5\%   & 78.8\%  & 13.7\%  \\ \hline
Image Diversity        & 21.3\%  & 75.0\%  & 3.7\%   \\ \hline
Story Engagement       & 37.5\%  & 55.0\%  & 7.5\%   \\ \hline
Image-Story Coherence  & 5.0\%   & 86.3\%  & 8.7\%   \\ \hline
\end{tabular}
\caption{The results of user study between MM-Interleaved, SEED-Story, and Ties}
\label{tab:user_study}
\end{table}

Participants were asked to choose their preference based on image quality, image style consistency, image diversity, story engagement, and image-story coherence. The results were obtained from 125 samples, as shown in Table~\ref{tab:user_study}.

The results indicate that SEED-Story clearly outperforms the baseline in image generation ability and text-to-image coherence. Additionally, SEED-Story shows a slightly higher preference in text quality for story generation.

\section{Ablation study on de-tokenizer adaptation}
\label{sec:ablation_detokenizer_adaptation}
\begin{figure}[h]
\centering
\includegraphics[width=1.\linewidth]{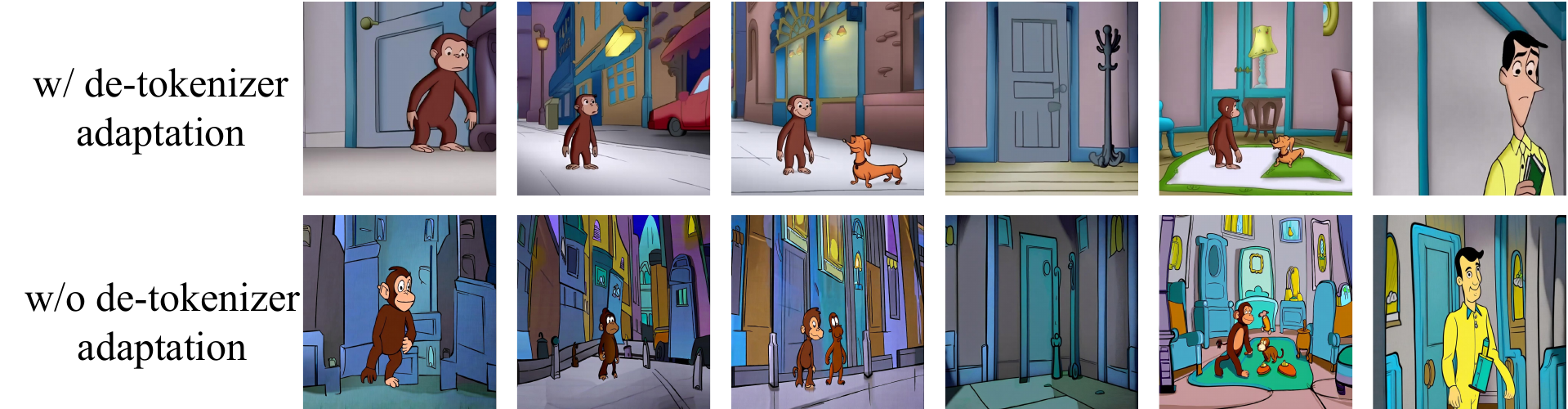}
\caption{Generated story images with and without the 3rd stage: \textbf{de-tokenizer adaptation}. The images generated before the third stage preserve semantic information, with mostly correct backgrounds and characters. However, they display low-quality textures and inconsistency in cartoon style. Our third stage effectively enhances these aspects.}
\label{fig:detokenizer}
\end{figure}

We find that the generated images before the de-tokenizer adaptation stage exhibit semantic relevance with consistent backgrounds and characters, thanks to MLLM's context preservation, as shown in Figure~\ref{fig:detokenizer}. However, they suffer from texture distortion and inconsistency in style. After de-tokenizer adaptation, the images show improved consistency in style and character appearance. The calculated FID scores in table~\ref{tab:detokenizer_ablation} confirm that de-tokenizer adaptation enhances image quality.

\begin{table}[ht]
\centering
\begin{tabular}{l|c}
\hline
\textbf{Model} & \textbf{FID} \\ \hline
Before 3rd stage & 153.93 \\ \hline
After 3rd stage  & 99.79  \\ \hline
\end{tabular}
\caption{FID scores before and after the 3rd stage.}
\label{tab:detokenizer_ablation}
\end{table}

\section{Implementation Details}
\subsection{Visual Tokenization and De-tokenization}
For visual tokenization, we use Qwen-VL pre-trained ViT-G. We first resize the image to 448x448 images and then use ViT to produce its feature of length256 with 4096 dimension. (shape: [256, 4096]). Inside the MLLM, we use a Q-Former architecture to process the image embedding. It takse the ViT image feature as key and Value, and conduct attention with its learnable queries. The length of learnable queries are 64.
For de-tokenization, we also use a Q-Former architecture to transform the MLLM output to the shape of SD-XL condition embedding.

\subsection{Instruction Tuning}
Instruction tuning data is formatted as follows: for each story, we sample a random length and compute losses on the last sequence (highlighted in red text). The sequence format is structured as:

\begin{quote}
    \texttt{<bos>[start of the story.][User prompt: ][following sequence 1][following sequence 2][following sequence 3][following sequence 4] ... \textcolor{red}{[target sequence]<eos>}}
\end{quote}

For our language model (LLM), we utilize the LLAMA2-7B pre-trained model and finetune it using LoRA, supported by the \textit{peft} library. The hyperparameter $r$ is set to 6, and $\text{lora\_alpha}$ is set to 32. The modules optimized include the $q\_projection\_layer$, $v\_projection\_layer$, $k\_projection\_layer$, $o\_projection\_layer$, $gate\_projection\_layer$, $down\_projection\_layer$, and $up\_projection\_layer$. We employ a learning rate of $1 \times 10^{-4}$ to finetune this model on our dataset across approximately 6 epochs, utilizing 8 NVIDIA-A800 GPUs.

\subsection{De-tokenizer Adaptation}
In this stage we fully finetune the SD-XL model. The data format is as the same as instruction tuning, but we fix all MLLM params and optimize only the SD-XL. It takes the MLLM output and is asked to produce image correspond to the ground truth. The SD-XL model was trained using 4 NVIDIA-A800 GPUs. A learning rate of $1 \times 10^{-4}$ was chosen to facilitate gradual weight updates, ensuring stable convergence, while a weight decay of 0.03 was applied for regularization to prevent overfitting. Training was performed using mixed precision (\texttt{bf16}), which significantly reduced memory usage and accelerated the training process without compromising the model's accuracy. The model underwent three training epochs, balancing the learning of complex patterns against computational resource use, optimized for large-scale datasets and sophisticated model architectures.

\section{Analysis of Multimodal Attention Sink}
\label{sec:appendix_attn_sink}
\subsection{Attention Map Visualization}
In this section, we present additional visualizations of attention maps. These maps are derived from various model runs, including varying data lengths, attention heads, and layers. The visualizations consistently reveal a pattern of attention focused on ``0'' tokens, punctuation, tokens adjacent to Begin-of-Image (BoI), and tokens adjacent to End-of-Image (EoI).

\begin{figure}[h]
% \begin{adjustwidth}{-0.05\linewidth}{-0.05\linewidth}
\centering
\includegraphics[width=1\linewidth]{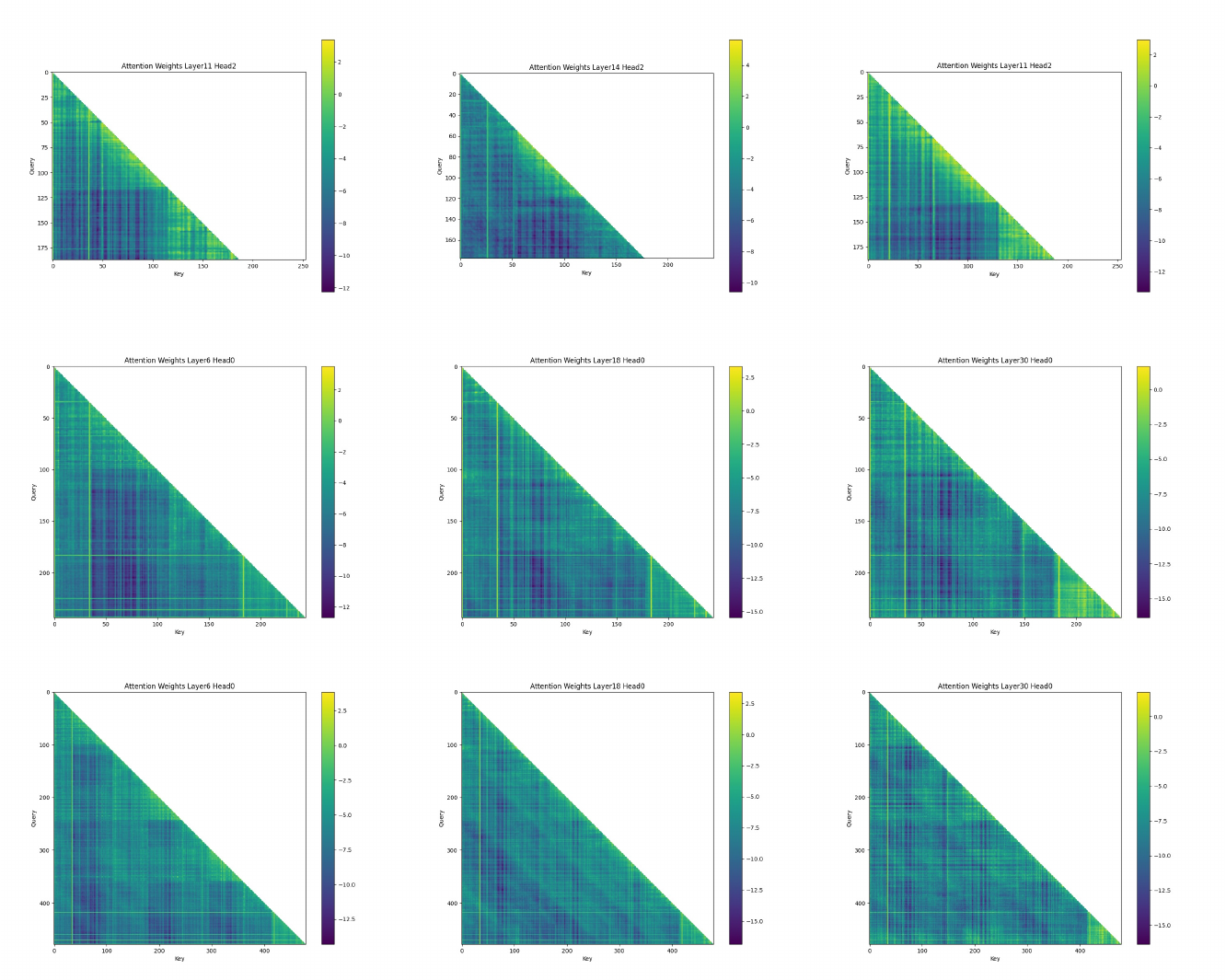}
\caption{Visualization of attention maps from various model runs, showcasing attention patterns across different data lengths, attention heads, and model layers. Notably, the maps highlight consistent focus on '0' tokens, punctuation, tokens adjacent to Begin-of-Image (BoI), and tokens adjacent to End-of-Image (EoI).}
\label{fig:sup_attn_map}
% \end{adjustwidth}
\end{figure}

\subsection{Statistics on Token's Attention}
To demonstrate that more attention is paid to image tokens adjacent to \textbf{<BOI>} and \textbf{<EOI>}, we analyzed over 5600 attention maps from various models, layers, and input sequences, to identify ``key tokens'' with the highest mean attention values. For each attention map, we computed the mean attention value across the query dimension for every key and selected the top 10 keys. We then aggregated these results to determine how often each token appeared as a key token. Table~\ref{tab:token_occurrence} shows the tokens with the highest occurrences, supporting our multimodal attention sink mechanism, with most queries focusing on four key token categories.
\begin{itemize}
    \item Starting tokens (BOS)
    \item Punctuation (``,'' ``.'' and ``;'' ...)
    \item Image tokens near BOI (BOI, IMG04)
    \item Image tokens near EOI (from IMG57 to IMG63, added EOI)
\end{itemize}

\begin{table}[ht]
\centering
\begin{tabular}{l|c|c|c|c|c|c}
\hline
\textbf{} & \textbf{BOS} & \textbf{IMG57} & \textbf{EOI} & \textbf{IMG04} & \textbf{,} & \textbf{IMG60} \\ \hline
\textbf{Key Token Occurrence} & 4320 & 4320 & 4320 & 4140 & 4140 & 4120 \\ \hline
\textbf{} & \textbf{IMG61} & \textbf{IMG59} & \textbf{IMG62} & \textbf{IMG63} & \textbf{BOI} & \textbf{IMG56} \\ \hline
\textbf{Key Token Occurrence} & 3730 & 3132 & 1651 & 607 & 603 & 361 \\ \hline
\end{tabular}
\caption{Key Token Occurrence for Various Tokens}
\label{tab:token_occurrence}
\end{table}

\section{Story Visualization Comparison}\label{sec:story_visualization}

\begin{wraptable}{r}{0.4\textwidth} % This aligns the table to the right and sets it to half the text width
  \centering
  \vspace{-0.2in}
  \caption{Quantitative evaluation for story visualization.}
  \begin{tabular}{lcc}
    \hline
    \textbf{Model} & \textbf{FID $\downarrow$} & \textbf{CLIP Score$\uparrow$}  \\ \hline
    LDM & 67.29 & 0.7585 \\
    StoryGen & 73.74 & 0.7573 \\
    \textbf{SEED-Story} & \textbf{67.01} & \textbf{0.7793} \\ \hline
  \end{tabular}
  \label{tab:story_vis}
  \vspace{-0.1in}
\end{wraptable}

\begin{figure}[h]
% \begin{adjustwidth}{-0.05\linewidth}{-0.05\linewidth}
\centering
\includegraphics[width=1\linewidth]{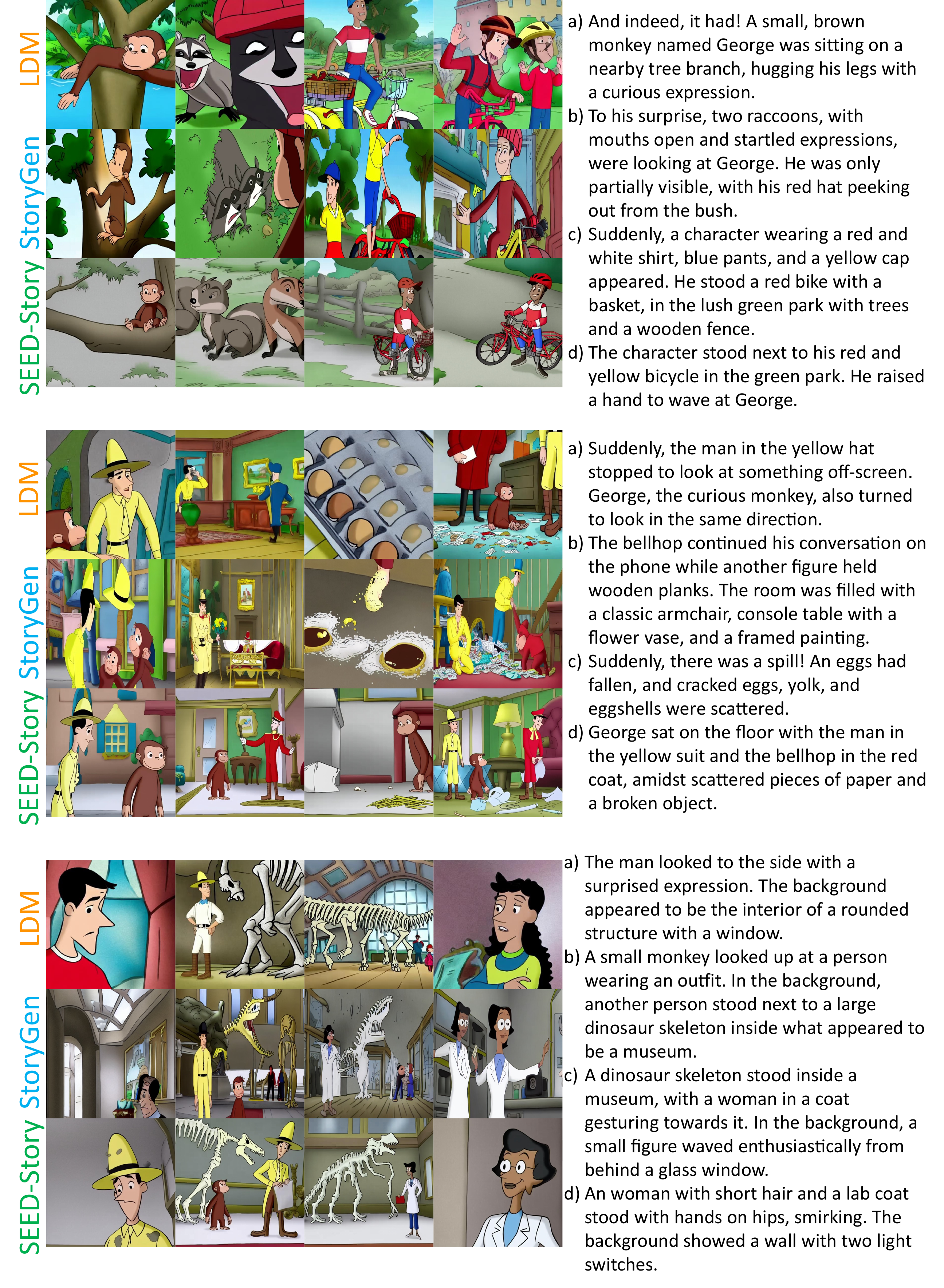}
\caption{Story visualization comparison of SEED-Story and other story visualization methods.}
\label{fig:story_vis_sup}
% \end{adjustwidth}
\end{figure}

\begin{figure}[h]
% \begin{adjustwidth}{-0.05\linewidth}{-0.05\linewidth}
\centering
\includegraphics[width=1\linewidth]{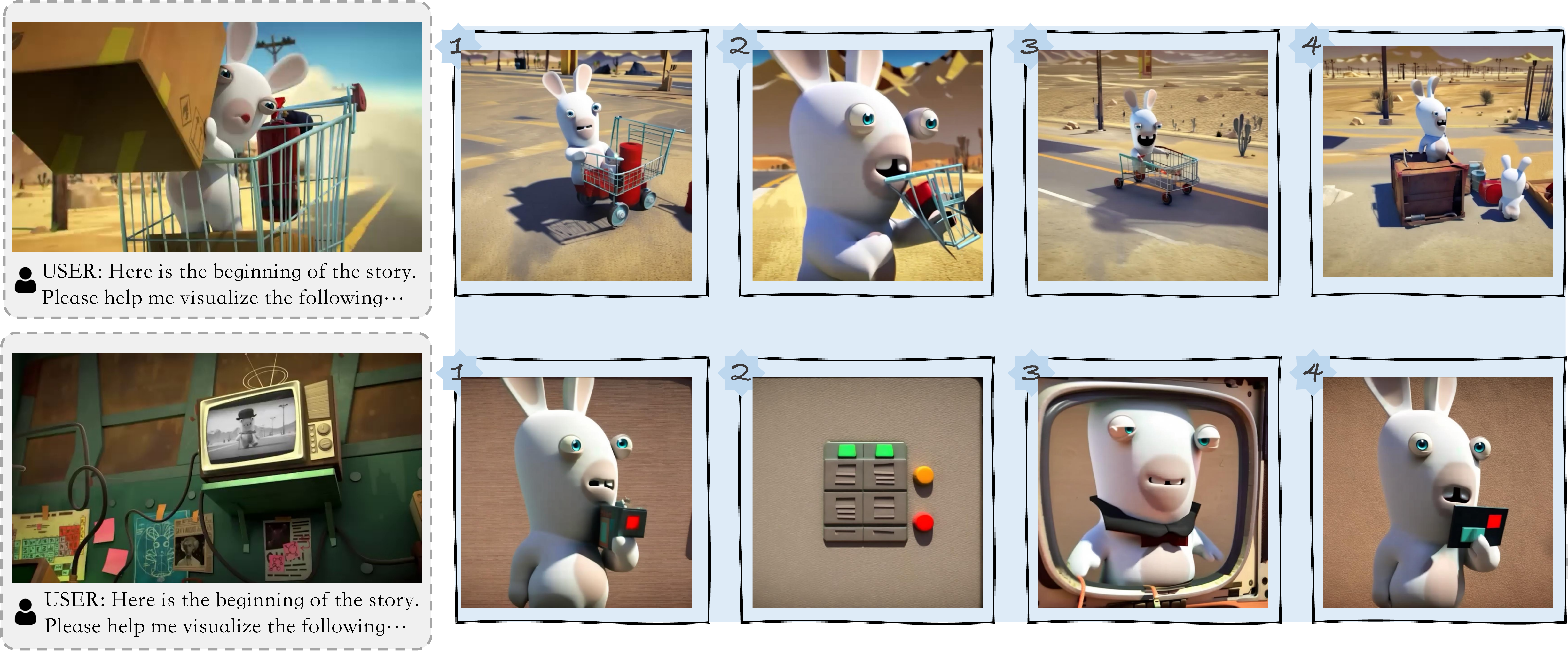}
\caption{Story visualization result on Rabbids Invasion data.}
\label{fig:sup_gen_rabbids}
% \end{adjustwidth}
\end{figure}

\begin{figure}[h]
% \begin{adjustwidth}{-0.05\linewidth}{-0.05\linewidth}
\centering
\includegraphics[width=1\linewidth]{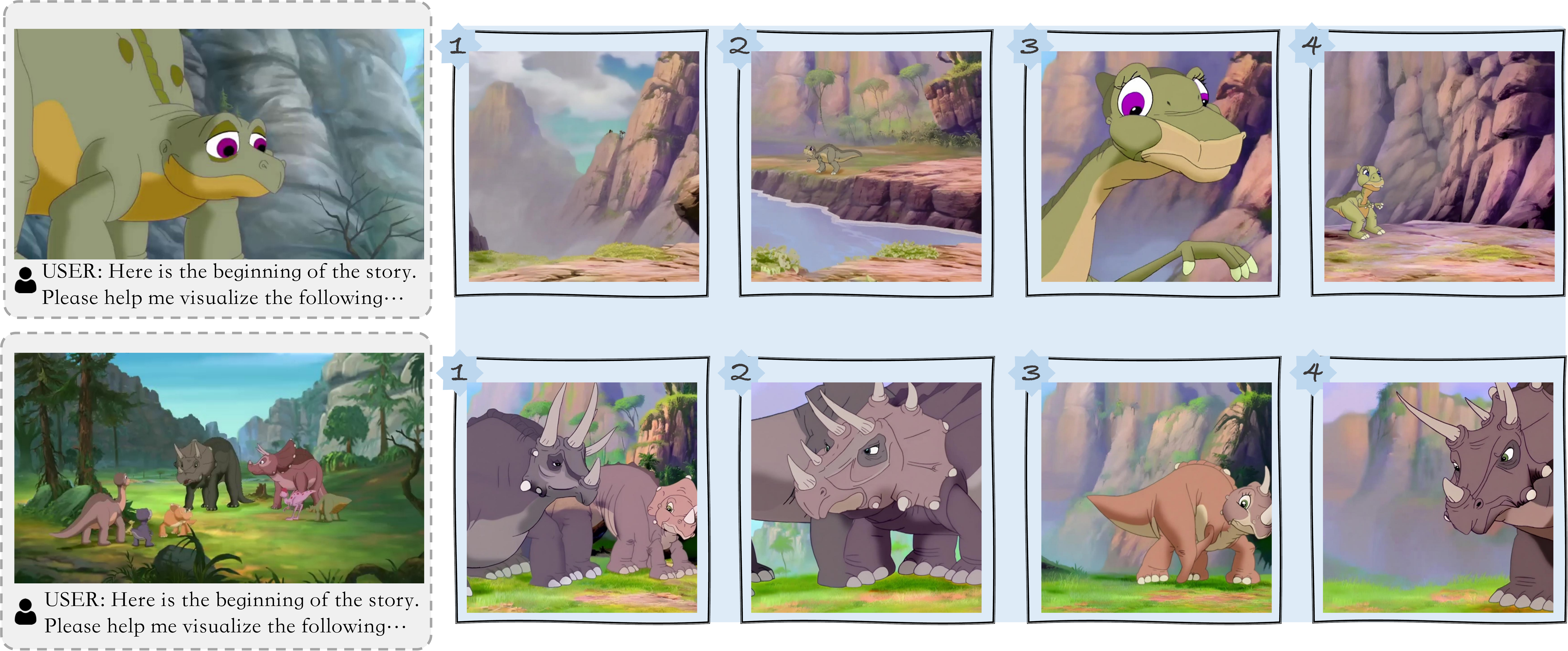}
\caption{Story visualization result on The Land Before Time data.}
\label{fig:sup_gen_the_land}
% \end{adjustwidth}
\end{figure}

% \vspace{-0.1in}
Previous story generation approaches primarily utilize diffusion models, focusing on visualizing story images. These models take the previous image and text as input, and then generate only the next image based on the current text prompt. For a fair comparison, we adapt our model to a visualization-only format. For StoryGen~\cite{liu2023intelligent}, we also train it to produce images with previous images and texts. For LDM~\cite{rombach2022high}, we only give it text-image pairs. The visual results are shown in Figure~\ref{fig:story_vis_sup}. SEED-Story model shows better style and character consistency and higher quality compared to baselines. We also showcase the visualization result of our model on Rabbids Invasion and The Land Before Time. Please see Figure~\ref{fig:sup_gen_rabbids} and \ref{fig:sup_gen_the_land}. We further conduct a quantitative evaluation in Table~\ref{tab:story_vis} to demonstrate our effectiveness.

%In this section, we present additional visualization comparison of our SEED-Story and other story visualization methods. SEED-Story shows better image consistency and higher qualit, as shown in Figure~\ref{fig:story_vis_sup}.
\iffalse
\begin{figure}[h]
% \begin{adjustwidth}{-0.05\linewidth}{-0.05\linewidth}
\centering
\includegraphics[width=1.0\linewidth]{figs/story_vis_less_seed.pdf}
\vspace{-0.35in}
\caption{Story visualization comparison between SEED-Story and baseline models. SEED-Story generates images with higher quality and better consistency.}
\label{fig:story_vis}
% \end{adjustwidth}
\end{figure}
\fi

% \vspace{-0.1in}

\section{Multimodal Story Generation Results}
\label{sec:visual_results}
\begin{figure}[p]
\centering
\includegraphics[width=1.\linewidth]{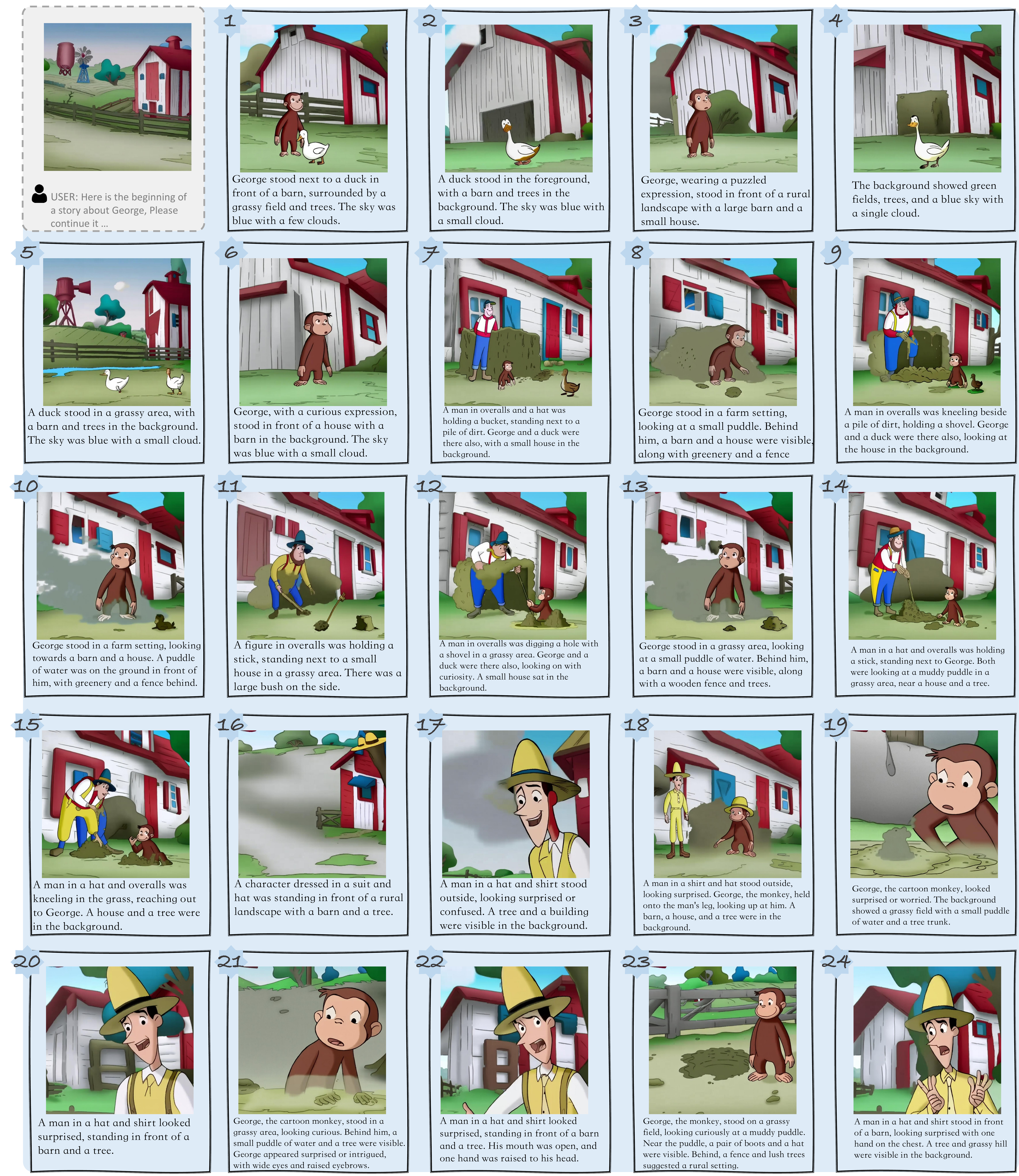}
\caption{Multimodal long story generation results of SEED-Story.}
\label{fig:multimodal_story_gen_1}
\end{figure}

\begin{figure}[p]
\centering
\includegraphics[width=1.\linewidth]{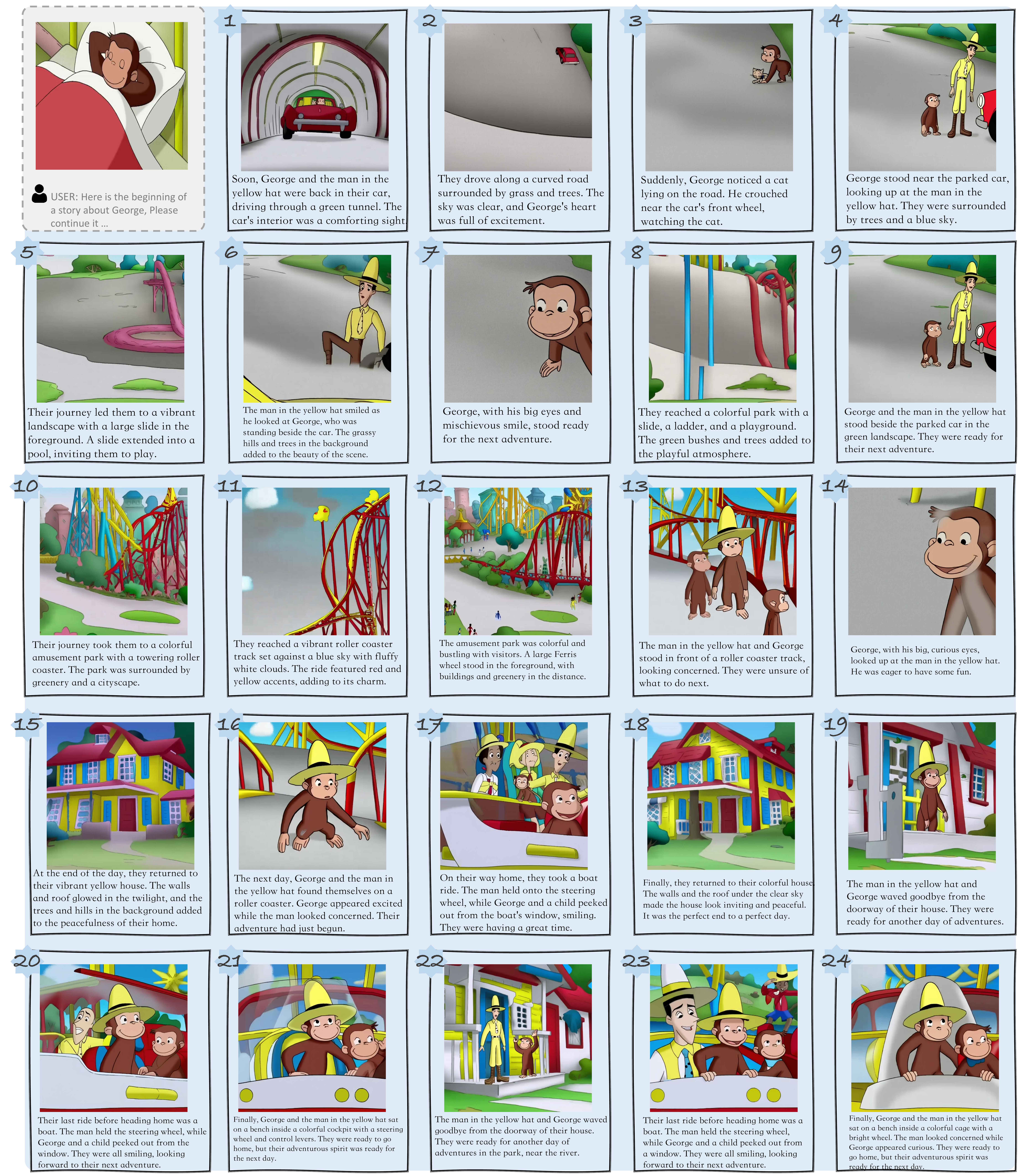}
\caption{Multimodal long story generation results of SEED-Story.}
\label{fig:multimodal_story_gen_2}
\end{figure}

In this section, we present more multimodal story generation results of our SEED-Story. It keeps produce story image and text with high quality.
Figure~\ref{fig:multimodal_story_gen_1}, Figure~\ref{fig:multimodal_story_gen_2}, and Figure~\ref{fig:sup_gen_case3} prove our multimodal long story generation capabilities. SEED-story can generate long sequences with engaging plots and vivid images.

\iffalse
\vspace{0.3in}
\begin{figure}[h]
\centering
\includegraphics[width=.95\linewidth]{figs/multi_gen_case.pdf}
\caption{Examples of multimodal story generation from SEED-Story. It shows two narrative branches generated from the same initial image. The top branch starts with text referencing ``the man in the yellow hat,'' leading to images that include the character. The bottom branch starts without mentioning the man, resulting in stories that diverge from the first by excluding him.}
\label{fig:multimodal_story_gen_control}
\end{figure}

\vspace{0.3in}
\fi

\begin{figure}[h]
% \begin{adjustwidth}{-0.05\linewidth}{-0.05\linewidth}
\centering
\includegraphics[width=1\linewidth]{figs/sup_gen_case3.pdf}
\caption{Multimodal story generation results of SEED-Story.}
\label{fig:sup_gen_case3}
% \end{adjustwidth}
\end{figure}

% \begin{figure}[h]
% % \begin{adjustwidth}{-0.05\linewidth}{-0.05\linewidth}
% \centering
% \includegraphics[width=1\linewidth]{figs/sup_gen_case2.pdf}
% \caption{Multimodal story generation results of SEED-Story.}
% \label{fig:sup_gen_case2}
% % \end{adjustwidth}
% \end{figure}

% \begin{figure}[h!]
% % \begin{adjustwidth}{-0.05\linewidth}{-0.05\linewidth}
% \centering
% \includegraphics[width=1\linewidth]{figs/sup_gen_case1.pdf}
% \caption{Multimodal story generation results of SEED-Story.}
% \label{fig:sup_gen_case1}
% % \end{adjustwidth}
% \end{figure}

\section{Details about GPT-4V Evaluation}~\label{sec:evaluation}
\subsection{Comparative Evaluation}
To evaluate the effectiveness of MM-interleaved and SEED-Story in multimodal story generation, we initiate an experiment where each model produces a story of five segments, based on a common starting image and text. The segment limit is set to five to accommodate the constraints of GPT-4V, which can handle a maximum of ten images per input session. In total, we generate 180 stories for assessment. For evaluation, we employ GPT-4 or GPT-4V to determine which model produces the better story in each case, based on the framework established in L-Eval~\cite{an2023leval}. We calculate the win rate for each model to determine its performance relative to its counterpart. The prompt we used is shown below.

\begin{quote}
    ``Please act as an impartial judge and evaluate the quality of the generation story contents provided by two AI assistants. Your job is to evaluate which assistant's generation is better. Your evaluation should consider \textbf{\{the style consistency of the story images / the engagement of the story / the coherence of the generated text and images\}}. Avoid any position biases and ensure that the order in which the responses were presented does not influence your decision. Do not allow the length of the responses to influence your evaluation. Do not favor certain names of the assistants.Be as objective as possible. After providing your explanation, output your final verdict by strictly following this format: ``[[A]]'' if assistant A is better, ``[[B]]'' if assistant B is better, and ``[[C]]'' for a tie.''
\end{quote}

\subsection{Score Evaluation}
We also provide a prompt for directly estimating the performance of the generated results without comparing to others. The prompt we used is shown below. We present the direct estimation score is shown in Table~\ref{tab:score_eval}

\begin{quote}
    ``Please act as an impartial judge and evaluate the quality of the generation story contents provided by an AI assistant. Your job is to give a score out of 10. Your evaluation should consider \textbf{\{the style consistency of the story images / the engagement of the story / the coherence of the generated text and images\}}. Do not allow the length of the responses to influence your evaluation. Be as objective as possible. After providing your explanation, output your final score by strictly following this format: ``[[score]]'', such as ``[[7]]''.''
\end{quote}

\begin{table}[h]
  \centering
  \caption{GPT4 score evaluation results in 3 different aspects-style consistency, story engaging level, and text-image coherence.}
  \begin{tabular}{lccc}
    \hline
     & \textbf{Style $\uparrow$} & \textbf{Engaging$\uparrow$}  & \textbf{Coherence$\uparrow$}\\ \hline
    \textbf{SEED-Story} & 8.61 & 6.27 & 8.24 \\ \hline
  \end{tabular}
  \label{tab:score_eval}
\end{table}

\section{Story Video}
To showcase the capabilities of our multimodal generation model, we employ a video generation technique to animate the images. We then synchronize these moving images with audio to create a narrative video, which is available in our supplementary materials.

\section{Data Usage and License}
\subsection{Curious George}
Curious George is an animated series featuring George, a curious monkey whose adventures teach preschoolers about math, science, and engineering. Guided by The Man with the Yellow Hat, George explores the world through problem-solving and experimentation, making it a delightful and educational experience for young viewers.

Curious George is released on PBS KIDS~\cite{CuriousGeorgePBSKids, CuriousGeorgeYouTube}, a not-for-profit institution. It is a production of Imagine, WGBH and Universal. Curious George and related characters, created by Margret and H.A. Rey, are copyrighted and trademarked by Houghton Mifflin Harcourt and used under license. Licensed by Universal Studios Licensing LLC. Television Series: ©2024 Universal Studios. The terms of use of them are provided in \url{https://www.pbs.org/about/about-pbs/terms-of-use/}. 

Our usage fully comply with the terms of use. 1) Personal Uses Permitted: My project is non-commercial and educational, which aligns with personal uses as outlined by PBS. we are not using the information for commercial purposes or exploiting it in a manner inconsistent with PBS rules. The use is strictly for educational and research purposes within an academic setting.
2) User's Obligation to Abide By Applicable Law: We will ensure all research activities comply with local laws, particularly those relating to copyright and intellectual property rights. Our use will not involve unauthorized reproduction, distribution, or exhibition that violates Intellectual Property Laws. All data are for research only.
3) Content of Information: We will responsibly use the "Curious George" materials, ensuring that all content used in our research is accurately cited and acknowledged. Any PBS content incorporated into your project will be clearly attributed to PBS.

\subsection{Rabbids Invasion}
``Rabbids Invasion'' is a French-American computer-animated TV series that breathes life into the zany antics of Ubisoft's popular Rabbids video game characters. Created by Jean-Louis Momus and featuring the voice of Damien Laquet, the show is a dynamic blend of humor and adventure tailored for a family audience. Since its debut on August 3, 2013, on France 3, the series has enjoyed multiple seasons and a global reach. The Rabbids are mischievous rabbit-like creatures whose escapades lead them into all sorts of unpredictable and hilarious situations, making ``Rabbids Invasion'' a delight for both kids and adults alike. Thanks to their release, we derive some subsets from the cartoon series Rabbids Invasion~\cite{RabbidsInvasionYouTube, Animaj}.

\subsection{The Land Before Time}
The Land Before Time, an iconic animated film series created by Judy Freudberg and Tony Geiss and distributed by Universal Pictures, debuted in 1988 with significant contributions from Don Bluth, George Lucas, and Steven Spielberg. This franchise, consisting of an initial film followed by 13 sequels, a TV series, video games, and extensive merchandising, explores the adventures of five young dinosaurs who learn key life lessons about friendship and teamwork through their prehistoric trials. Despite the absence of the original creators in the sequels, the series has continued to captivate audiences, emphasizing themes of community and perseverance across its extensive narrative arc. Thanks to their release, we derive some subsets from their websites~\cite{TheLandBeforeTime, TheLandBeforeTimeWebsite}. 

\subsection{Appreciation}
Leveraging the data derived from "Curious George," "Rabbids Invasion," and "The Land Before Time," we have significantly advanced the capabilities of our story generation models. This progress has direct and impactful implications for children's education by enhancing their imaginative faculties and fostering a keen interest in learning. By integrating elements from these animated series into our models, we not only engage young minds but also deepen their affection for animated storytelling. Consequently, this not only meets but also amplifies educational objectives, such as improving literacy and cognitive skills through enjoyable and interactive content. The successful application of data from these beloved animations in our research exemplifies how academic pursuits can harmoniously blend with educational entertainment, ultimately delivering multifaceted benefits that extend well beyond conventional learning environments.

Lastly, we extend our profound appreciation to the creators and maintainers of "Curious George," "Rabbids Invasion," and "The Land Before Time," each a rich and vibrant resource that has significantly contributed to the scope and success of our research. The engaging narratives and characters from these series, especially the ever-curious George, the mischievous Rabbids, and the adventurous dinosaurs from The Land Before Time, have provided invaluable data that enhanced our narrative generation models. This project benefited immensely from the educational and entertaining content crafted with meticulous attention to detail, fostering imagination and learning in young audiences. We acknowledge the pivotal role that these animated series have played in advancing academic research aimed at educational technology. The commitment of the teams behind these beloved series to fostering curiosity and learning is both inspiring and exemplary. We are immensely grateful for the opportunity to incorporate such cherished resources into our scholarly work.

\section{Broader Impacts}
This project may potentially produce copyrighted content, particularly when used inappropriately or without adherence to existing intellectual property laws. To mitigate this risk, we will implement a rigorous compliance framework that respects the copyrights of third parties. This involves setting strict usage licenses that align with the legal standards dictated by our data sources. Our aim is to protect intellectual property rights while fostering innovation and ethical use of our technology.
We also commit to educating users on the importance of respecting intellectual property rights when using our technology. This will be achieved through detailed user guidelines, training sessions, and readily available support to help users understand and navigate the complexities of copyright laws.
By taking these measures, we aim not only to comply with legal standards but also to promote a culture of respect for intellectual property within our user community, thereby contributing positively to the broader digital ecosystem.

\section{Limitations}
\paragraph{Lack of Realistic Data Experimentation:} This limitation points to a potential gap in the validation of the SEED-Story model under practical, real-world conditions. Without experiments using realistic data, it's difficult to ascertain how the model would perform in scenarios that are not perfectly controlled or that deviate from the training conditions. This can be crucial, especially in applications like storytelling where the context and variability of real-world data play significant roles. A possible solution would be to incorporate a broader range of test conditions, including noisy data or data from "in-the-wild" storytelling scenarios, to evaluate the robustness and adaptability of the model.

\paragraph{Training on a Non-Diverse Dataset:} The second limitation is the restriction of the model's training to animation datasets which does not cover a large scale or diverse styles. This can severely limit the model's ability to generalize and produce outputs in styles that are not represented in the training data. This is particularly limiting in creative tasks such as storytelling where the ability to adapt to various artistic and narrative styles is crucial. To mitigate this, expanding the dataset to include a wider array of styles, genres, and visual aesthetics could be beneficial.

\clearpage

\end{document}